\documentclass[10pt,twocolumn,letterpaper]{article}

\usepackage{cvpr}              
\usepackage{algorithm}
\usepackage{algorithmic}
\usepackage{amsfonts}
\usepackage{amsmath}
\usepackage{subcaption}
\usepackage{tabularx} 
\usepackage{caption}

\usepackage{multirow} %
\usepackage{booktabs}  %
\usepackage{listings}
\usepackage{pifont}
\definecolor{myblue}{RGB}{220,230,242}  
\definecolor{mygreen}{RGB}{226,240,217} 

\usepackage[table]{xcolor}
\definecolor{cvprblue}{rgb}{0.21,0.49,0.74}
\usepackage[pagebackref,breaklinks,colorlinks,allcolors=cvprblue]{hyperref}

\def\confName{CVPR}
\def\confYear{2026}

\title{CASL: Concept-Aligned Sparse Latents for Interpreting Diffusion Models}
\author{
Zhenghao He, Guangzhi Xiong, Boyang Wang, Sanchit Sinha, Aidong Zhang \\
University of Virginia, USA \\
{\tt\small \{zhenghao, guangzhi, usy5km, sanchit, aidong\}@virginia.edu}
}

\begin{document}
\maketitle
\begin{abstract}
Internal activations of diffusion models encode rich semantic information, but interpreting such representations remains challenging. While Sparse Autoencoders (SAEs) have shown promise in disentangling latent representations, existing SAE-based methods for diffusion model understanding rely on unsupervised approaches that fail to align sparse features with human-understandable concepts. This limits their ability to provide reliable semantic control over generated images. 
We introduce \textbf{CASL} (Concept-Aligned Sparse Latents), a supervised framework that aligns sparse latent dimensions of diffusion models with semantic concepts. CASL first trains an SAE on frozen U-Net activations to obtain disentangled latent representations, and then learns a lightweight linear mapping that associates each concept with a small set of relevant latent dimensions.
To validate the semantic meaning of these aligned directions, we propose \textbf{CASL-Steer}, a controlled latent intervention that shifts activations along the learned concept axis. Unlike editing methods, CASL-Steer is used solely as a causal probe to reveal how concept-aligned latents influence generated content.
We further introduce the \textbf{Editing Precision Ratio (EPR)}, a metric that jointly measures concept specificity and the preservation of unrelated attributes. 
Experiments show that our method achieves superior editing precision and interpretability compared to existing approaches. To the best of our knowledge, this is the first work to achieve supervised alignment between latent representations and semantic concepts in diffusion models.
\end{abstract}    
\section{Introduction}
\label{sec:intro}

Diffusion models have achieved remarkable success in image generation. Recent studies~\cite{kwon2022diffusion, zhu2023boundary, jeong2024training, yang2024unleashing} have revealed that internal activations of diffusion models already exhibit rich semantic properties, with high-level concepts implicitly represented within their feature spaces.
In the field of natural language processing, Sparse Autoencoders (SAEs) have proven effective in disentangling latent semantics from internal activations~\cite{cunningham2023sparse}. Recent works have begun to adapt this approach to diffusion models, aiming to extract interpretable features from hidden activations during generation~\cite{yu2025ttfdiffusion, huang2025tide, Ijishakin, surkov2025unpacking}. These efforts show promises in revealing meaningful structures within the high-dimensional hidden states of diffusion models.

However, extending sparse interpretability to vision diffusion models remains challenging.
Existing attempts rely on unsupervised SAEs that attribute semantics to each latent unit by identifying the input that maximally activates it.
While this heuristic is effective in NLP, where discrete tokens naturally serve as semantic anchors, it breaks down in vision settings.
Image representations are continuous, spatially distributed, and heavily entangled, meaning that high activations often reflect a mixture of correlated attributes rather than a single interpretable concept.
For instance, the notion of “smiling’’ co-occurs with changes in the mouth, cheeks, eyes, and global facial geometry, making it unclear which latent unit should be associated with the concept itself.
As a result, current vision SAEs struggle to produce consistent, concept-specific latent directions, limiting their ability to support meaningful interpretation or controlled intervention.


To overcome these limitations, we introduce CASL, a framework that learns concept-aligned sparse latents in diffusion models.
CASL first trains a Sparse Autoencoder on frozen U-Net activations to obtain a disentangled latent space.
Then, instead of relying on unsupervised heuristics, we incorporate weak supervision to learn a lightweight linear mapping that associates each concept with its most relevant sparse dimensions.
This yields a set of concept-aligned latent directions that are suitable for semantic interpretation.
To verify their semantic effect, we further apply CASL-Steer, a controlled latent intervention that perturbs activations along the aligned direction to evaluate how each concept influences the model’s generative behavior. 

Our main contributions are summarized as follows:
\begin{itemize}
\item We introduce \textbf{CASL}, a supervised framework that learns concept-aligned sparse latents in diffusion models. To our knowledge, this is the first method to explicitly align SAE latent dimensions in diffusion models with human-defined semantic concepts.

\item We propose \textbf{CASL-Steer}, a controlled latent intervention used to validate the semantic effects of the aligned latent directions. CASL-Steer serves as a causal probe for interpretability rather than an image editing method.

\item We design the \textbf{Editing Precision Ratio (EPR)}, a quantitative metric that jointly measures semantic specificity and the preservation of unrelated attributes, enabling systematic evaluation of concept alignment in sparse latent spaces.
\end{itemize}
\section{Related Work}
\label{sec:related}

\paragraph{Semantic Editing in Diffusion Models.}
Diffusion models have demonstrated impressive capabilities in semantic image editing by conditioning the generation process on user-defined signals. Early methods such as SDEdit~\cite{meng2021sdedit} apply noise to input images and guide the denoising trajectory via external conditioning, achieving structure-preserving edits. Prompt-based approaches like DiffusionCLIP~\cite{kim2022diffusionclip}, InstructPix2Pix~\cite{brooks2023instructpix2pix}, and Imagic~\cite{kawar2023imagic} steer generation using language or visual instructions, typically requiring fine-tuning of the diffusion model. In contrast, methods such as Prompt-to-Prompt~\cite{hertz2022prompt}, Asyrp~\cite{kwon2022diffusion}, and Concept Sliders~\cite{2024concept} explore direct intervention into internal representations of a frozen U-Net, enabling localized edits without retraining. Among them, Asyrp proposes lightweight, learnable residual modules inserted into intermediate layers to control semantic attributes. However, despite their effectiveness, these methods still operate as black boxes and provide little insight into how internal activations correspond to human-interpretable concepts—a gap our work seeks to address.

\paragraph{Sparse and Interpretable Representations.}  
Sparse Autoencoders (SAEs) have shown promise in learning human-interpretable representations by enforcing sparsity on latent activations. In natural language processing (NLP), SAEs trained on frozen transformer features have been shown to produce sparse latent dimensions that align with interpretable linguistic features, such as sentiment, negation, or syntactic structures~\cite{cunningham2023sparse, bricken2023towards}. This has inspired efforts to adapt SAE-based analysis to the vision domain, including generative models.

Recent studies have begun to explore SAE-based interpretation within diffusion models. Surkov et al.~\cite{surkov2025unpacking} and Ijishakin et al.~\cite{Ijishakin} both train SAEs on internal residual or bottleneck activations and demonstrate that manipulating individual latent units can causally affect generation. However, in the absence of semantic supervision, the learned features must be interpreted post-hoc and lack consistent alignment with human concepts. Huang et al.~\cite{huang2025tide} extend this paradigm to diffusion transformers, proposing a temporal-aware SAE architecture that captures denoising dynamics across timesteps. While their method improves feature disentanglement and supports classification, the latent space remains entangled with temporal and visual factors, and offers limited control over semantic content. Kim et al.~\cite{kim2024interpreting} adopt a more diagnostic perspective, using k-sparse autoencoders to analyze semantic granularity across layers and timesteps in various architectures. Although their analysis reveals monosemantic trends and enables transfer learning, it does not target controllable editing. 

Our approach introduces explicit concept-level supervision to directly align sparse latent units with human-interpretable concepts. This supervision ensures that specific latent features consistently reflect the presence or absence of target semantic attributes, enabling faithful and interpretable control over diffusion-based generation.

\begin{figure}[t]
\centering
\includegraphics[width=\columnwidth]{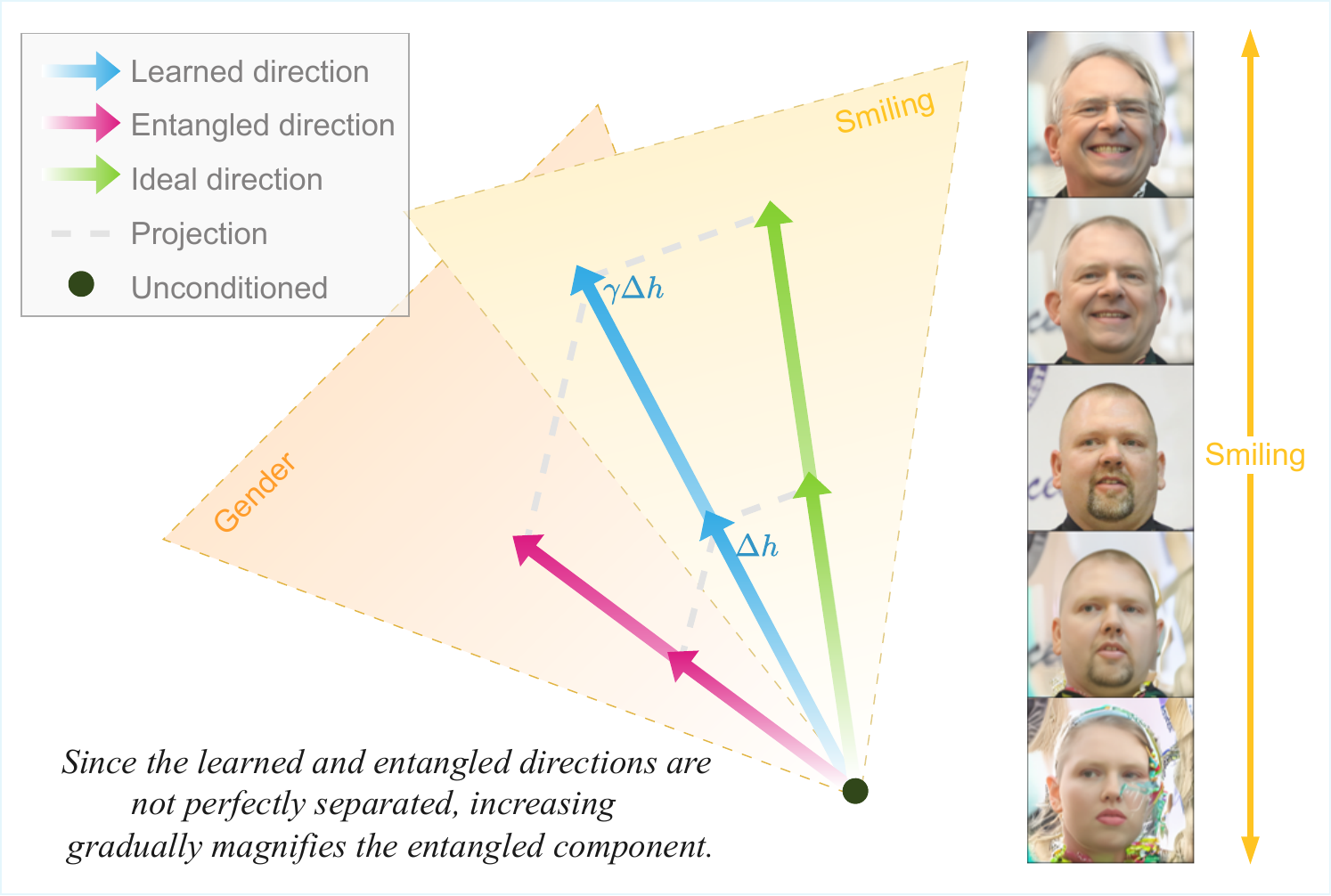}
\caption{Demonstration of semantic directions in the activation space. Left: Geometric illustration of ideal, entangled, and learned semantic directions in the activation space. Right: scaling the Asyrp direction \( \Delta h^{(\text{smile})} \) from \( \gamma = -2 \) to \( \gamma = 2 \) increases smiling intensity, but also introduces entangled changes in identity, hairstyle, and gender.}
\label{fig:problem}
\end{figure}

\begin{figure*}[t]

\centering
\includegraphics[width=\textwidth]{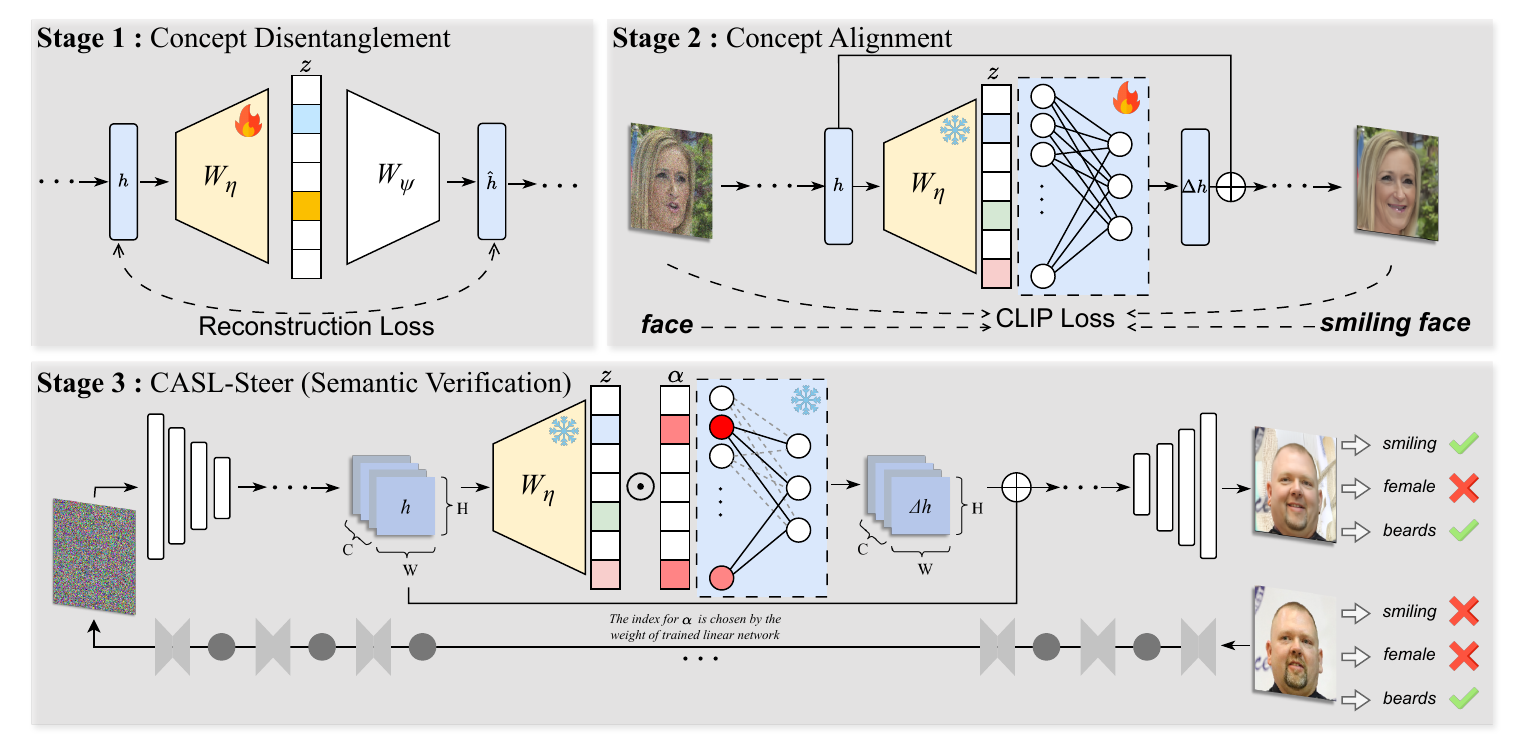}
\caption{
Overview of our proposed CASL framework. 
\textbf{Stage 1 (Concept Disentanglement):} 
A Sparse Autoencoder is trained on U-Net activations to obtain a structured sparse latent representation.
\textbf{Stage 2 (Concept Alignment):} 
A lightweight linear mapping aligns selected latent dimensions with human-defined semantic concepts, producing concept-aligned directions.
\textbf{Stage 3 (CASL-Steer):} 
A controlled latent intervention is applied along the aligned direction to \emph{verify} its semantic effect, serving as a probing mechanism.
}

\label{fig:framework}
\end{figure*}

\section{Problem Formulation}

\noindent{\textbf{U-Net Activations (\emph{h}-space).}}
Following ~\cite{kwon2022diffusion, zhu2023boundary, yu2025ttfdiffusion, Ijishakin}, we similarly denote the intermediate activation (\emph{h}-space) of a diffusion model's U-Net at a specific layer at $h \in \mathbb{R}^{C \times H \times W}$, where $C$, $H$, and $W$ denote the number of channels, height, and width, respectively. Prior works~\cite{zhu2023boundary, kwon2022diffusion} have shown that $h$ encodes rich semantic information and can be directly edited to control high-level concepts.

\noindent{\textbf{DDIM Sampling.}}
We adopt the DDIM sampling schedule~\cite{song2022denoisingdiffusionimplicitmodels} for deterministic denoising, which enables applying controlled interventions in \emph{h}-space without introducing stochastic variation.
Kwon et al.~\cite{kwon2022diffusion} introduces a learnable shift $\Delta h_t$ in the U-Net bottleneck, modifying only the $P_t$ term of the DDIM update:
\begin{equation}
    x_{t-1} = \sqrt{\alpha_{t-1}}\, P_t(\epsilon_\theta(x_t \mid \Delta h_t)) 
              + D_t(\epsilon_\theta(x_t)).
\end{equation}
This asymmetric activation-space injection enables controlled semantic manipulation without altering the overall generative trajectory. 

\noindent{\textbf{Structured Latent Hypothesis.}}
Empirical activation-space editing often exhibits entanglement: increasing the edit scale $\gamma_c$ strengthens the target attribute but simultaneously alters identity, background, or lighting. We argue that this arises because the learned direction $\Delta h^{(c)}$ is not a pure concept vector, but instead a mixture of multiple semantic components. Small edits suppress these mixed factors, while larger edits amplify them (see Fig.~\ref{fig:problem}).

To explain this behavior, we hypothesize that each concept direction $\Delta h^{(c)}$ admits a \emph{sparse expansion} over a shared set of latent semantic bases. Specifically, there exist basis elements
\(
\{b_i\}_{i=1}^K \subset \mathbb{R}^{C \times H \times W}
\)
such that
\begin{equation}
    \Delta h^{(c)}
    = \sum_{i=1}^{K} \beta^{(c)}_i\, b_i,
    \qquad
    \lVert \boldsymbol{\beta}^{(c)} \rVert_0 \ll K .
    \label{eq:latent_basis}
\end{equation}
Here, $\{b_i\}$ represent underlying semantic directions, and $\boldsymbol{\beta}^{(c)}$ encodes the concept’s coordinates in this basis. Our goal is to learn this structured latent space and align selected basis directions with human-defined concepts.

\section{Method}
\label{sec:method}

Our framework consists of three main stages: 
(1) We train a sparse autoencoder (SAE) to encode U-Net activations into a structured and interpretable latent space (Sec.~\ref{sec:stage1}). 
(2) We align selected latent dimensions with human-defined concepts by learning a lightweight linear mapping under semantic supervision (Sec.~\ref{sec:stage2}). 
(3) We verify the semantic effect of the aligned directions through a controlled latent intervention in activation space (Sec.~\ref{sec:stage3}).
In addition, we propose a metric to measure the effectiveness of aligning the latent representation with the semantic concepts (Sec.~\ref{sec:metric}). 
Fig.~\ref{fig:framework} illustrates the main components of our framework and their relationships.

\subsection{Concept Disentanglement (Stage 1)}
\label{sec:stage1}
Given an input activation $h \in \mathbb{R}^{C \times H \times W}$ from a diffusion model, we reshape $h$ into $N = H \times W$ tokens of dimension $C$, i.e., $h_{\text{seq}} \in \mathbb{R}^{N \times C}$. Following~\cite{huang2025tide, surkov2025unpacking}, each spatial location is treated as a token, and SAE maps $h_{\text{seq}}$ to sparse latent $z \in \mathbb{R}^{N \times K}$, where $K \gg C$.

The encoding and decoding processes are defined as:
\begin{align}
    z^{(t)} &= \phi \left( W_{\eta} (h^{(t)} + e(t) - b_{\mathrm{pre}}) + b_{\eta} \right) \\
    \hat{h}^{(t)} &= W_{\psi} z^{(t)} + b_{\mathrm{pre}}
\end{align}
where $h^{(t)} \in \mathbb{R}^{N \times C}$ is the U-Net activation at timestep $t$, and $z^{(t)} \in \mathbb{R}^{N \times K}$ is its sparse latent representation. The term $e(t) \in \mathbb{R}^{C}$ is a learnable timestep embedding added to $h^{(t)}$ before encoding. The encoder is a linear transformation $W_\eta \in \mathbb{R}^{K \times C}$ without bias, followed by a latent bias $b_\eta \in \mathbb{R}^K$ and an element-wise activation function $\phi(\cdot)$. The decoder is also linear, $W_\psi \in \mathbb{R}^{C \times K}$, without bias. A learnable pre-bias vector $b_{\mathrm{pre}} \in \mathbb{R}^C$ is subtracted from the input before encoding and added back after decoding to restore the original activation space.

The overall objective for training the SAE is:
\begin{equation}
\label{eq:sae_loss}
    \mathcal{L}_{\text{SAE}} = 
    \underbrace{\| h - \hat{h} \|_2^2}_{\text{Reconstruction Loss}} 
    + \underbrace{\lambda_{\text{sparse}} \| z \|_1}_{\text{Sparsity}}
\end{equation}
where $\lambda_{\text{sparse}}$ controls the strength of sparsity regularization.

\subsection{Concept Alignment (Stage 2)}
\label{sec:stage2}
After training the SAE, we freeze its encoder $W_\eta$ and perform inversion to obtain the disentangled latent representation $z = W_\eta(h)$ for each input activation $h$. 

To align specific latent dimensions $z_i$ with human-defined concepts (e.g., \emph{smiling}), we introduce a lightweight linear mapping without any activation function that predicts the concept editing direction in the activation space:
\begin{equation}
    \Delta h = W_{\Delta} z + b_{\Delta}
\end{equation}
where $W_{\Delta} \in \mathbb{R}^{C \times K}$ and $b_{\Delta} \in \mathbb{R}^C$ are learnable parameters.

The edited activation is computed as $h' = h + \Delta h$, which is injected into the U-Net to compute the modified noise prediction. We then use this modified output in the DDIM inversion formula to obtain the edited image:
\begin{equation}
    \hat{x}_0^{\text{edit}} = \frac{1}{\sqrt{\alpha_t}} \left( x_t - \sqrt{1 - \alpha_t} \cdot \varepsilon_\theta(x_t \mid h + \Delta h) \right)
\end{equation}

To ensure that $\Delta h$ effectively induces the desired concept change (e.g., adding a smile), we follow~\cite{kwon2022diffusion} and combine the DiffusionCLIP loss~\cite{kim2022diffusionclip} with an L1 regularization term:
\begin{align}
\mathcal{L} ={}& \lambda_{\mathrm{CLIP}} \, \mathcal{L}_{\mathrm{DiffusionCLIP}}(\hat{x}_0^{\text{edit}}, y^{\text{ref}}; x_0^{\text{origin}}, y^{\text{origin}}) \notag\\
& + \lambda_{\mathrm{recon}} \, \| \hat{x}_0^{\text{edit}} - x_0^{\text{origin}} \|_1
\label{eq:cliploss}
\end{align}
where $\hat{x}_0^{\text{edit}}$ denotes the generated image after applying the edited latent representation, and $x_0^{\text{origin}}$ is the input image without editing. $y^{\text{ref}}$ and $y^{\text{origin}}$ are the corresponding CLIP text embeddings of the target and original semantic descriptions, respectively.

Through this process, the model learns to associate certain latent dimensions in $z$ with explicit semantic concepts, enabling controllable and interpretable concept editing.





\begin{figure*}[t]
\centering
\includegraphics[width=\textwidth]{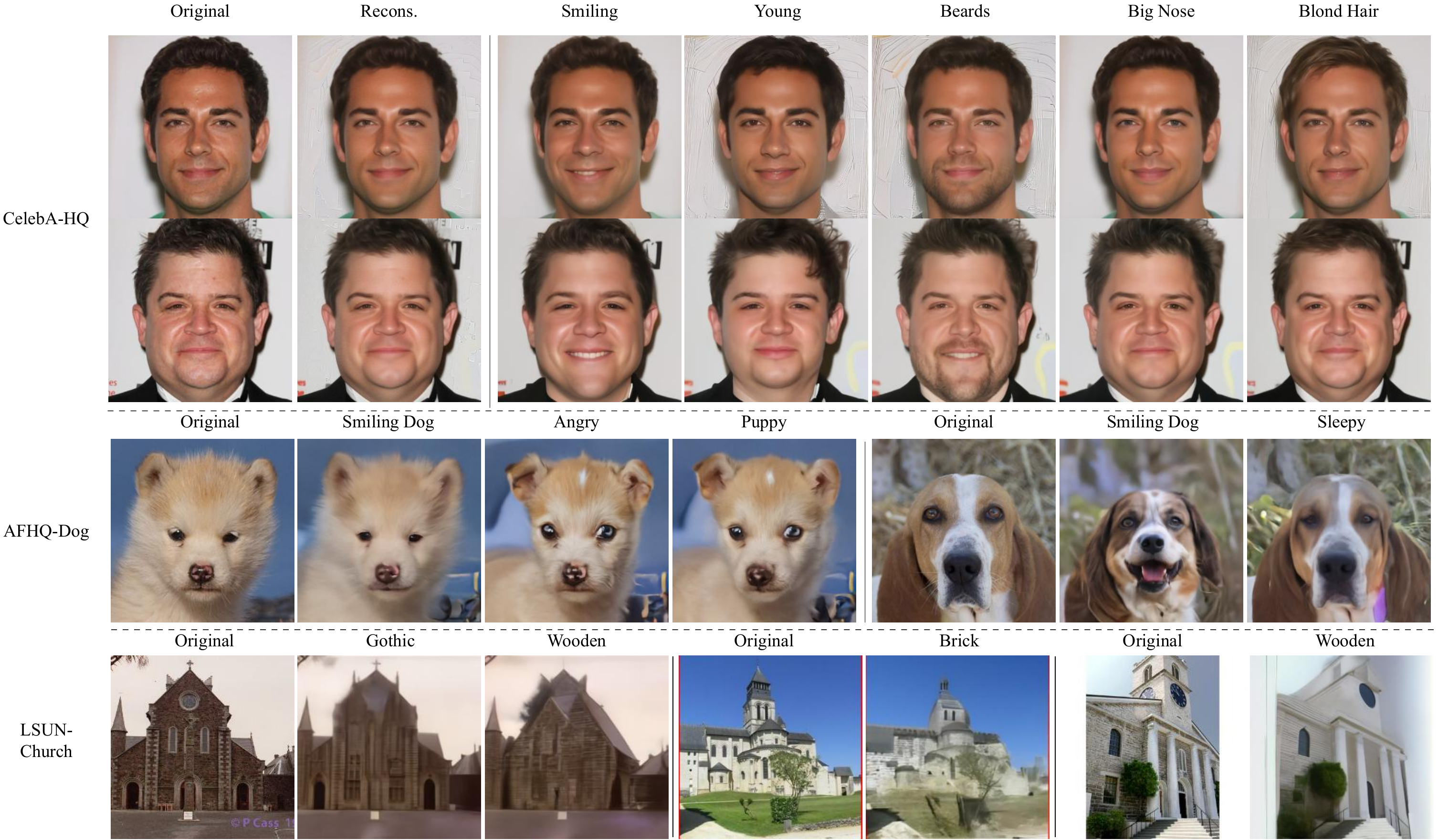}
\caption{\textbf{Attribute editing results of concept-aligned editing.} Our method enables concept-aligned editing across diverse attributes and domains. Each column shows a specific attribute (\textit{Smiling}, \textit{Young}, \textit{Puppy}, etc.) with consistent edits across different identities. Results are obtained by traversing sparse latent dimensions learned through supervised alignment.}
\label{fig:main_result}
\end{figure*}
\subsection{CASL-Steer: Semantic Verification (Stage 3)}
\label{sec:stage3}

Building on the concept alignment in Section~\ref{sec:stage2}, We aim to verify whether the learned sparse latent directions correspond to distinct, human-interpretable semantic concepts in the activation space, rather than focusing on image editing. To this end, we construct explicit mathematical formulations that enable principled and quantitative evaluation of concept alignment and disentanglement.

Firstly, we construct a sparse latent coordinate vector $\alpha \in \mathbb{R}^K$, where each entry specifies the traversal strength along a particular semantic direction. For a given concept $c$, we first select the top-$k$ latent dimensions as:
\begin{equation}
\label{eq:index_selection}
    \mathcal{I}_c = \operatorname{TopK}_k\left( \left| W_\Delta^{(c)} \right| \right)
\end{equation}
where $\left| W_\Delta^{(c)} \right|$ denotes the element-wise absolute value of the learned editing weights for concept $c$.

We then define the editing coordinate as
\begin{equation}
    \alpha_i = 
    \begin{cases}
        \alpha, & \text{if } i \in \mathcal{I}_c \\
        0, & \text{otherwise}
    \end{cases}
\end{equation}
where $\alpha$ controls the editing intensity. This construction ensures that editing is only applied along concept-specific, disentangled semantic directions.

Given a disentangled semantic basis $z$ for the current activation (obtained from the SAE encoder), the concept-wise editing direction in activation space is computed as:
\begin{equation}
\label{eq:s3_main}
    \Delta h_c =
    W_\Delta
        \big(
            \underbrace{\alpha}_{\substack{\text{coordinate in}\\\text{semantic basis}}}
            \odot
            \underbrace{z}_{\substack{\text{sparse}\\\text{semantic basis}}}
        \big)
\end{equation}
where $\odot$ denotes element-wise multiplication, $z$ represents the semantic basis vectors in the activation space, and $\alpha$ specifies the editing coordinate in this basis. The resulting activation shift \( \Delta h_c \) is then added to the original activation, and the edited image \( \hat{x}_0 \) is obtained following the same DDIM inversion process as described in Section~\ref{sec:stage2}.
\begin{table*}[ht]
\centering
\renewcommand{\arraystretch}{1.2}
\definecolor{mygray}{gray}{0.92}
\definecolor{lightgray}{gray}{0.4}
\newcommand{\pmgray}[1]{\textcolor{lightgray}{\text{\scriptsize$\pm$#1}}}

\resizebox{\textwidth}{!}{
\begin{tabular}{l l |c|ccccc|c}
\toprule
\multirow{2}{*}{\textbf{Concept}} &
\multirow{2}{*}{\textbf{Metric}} &
\multirow{2}{*}{\textbf{Recon.}} &
\multicolumn{5}{c|}{\textbf{Editing Methods}} &
\multicolumn{1}{c}{\textbf{Explainable}} \\
\cmidrule(lr){4-8} \cmidrule(lr){9-9}
& & &
\textbf{Boundary}~\cite{zhu2023boundary} &
\textbf{Asyrp}~\cite{kwon2022diffusion} &
\textbf{Slider}~\cite{2024concept} &
\textbf{MasaCtrl}~\cite{cao2023masactrl} &
\textbf{SwiftEdit}~\cite{Nguyen_2025_CVPR} &
\textbf{CASL-Steer} \\

\midrule
\multirow{4}{*}{Smiling}
& EPR~($\uparrow$) 
& 0.679\pmgray{0.289} 
& 1.231\pmgray{0.976} 
& 3.199\pmgray{2.182}
& 0.949\pmgray{0.479}
& 1.685\pmgray{1.375}
& \cellcolor{myblue}3.359\pmgray{2.064}
& \cellcolor{mygreen}\textbf{4.465}\pmgray{2.319} \\
& CLIP-Score~\cite{hessel2021clipscore}~($\uparrow$) 
& 0.126\pmgray{0.014}
& 0.166\pmgray{0.012}
& \cellcolor{myblue}0.192\pmgray{0.014}
& 0.178\pmgray{0.010}
& \cellcolor{mygreen}\textbf{0.204}\pmgray{0.013}
& 0.171\pmgray{0.015}
& \cellcolor{myblue}0.192\pmgray{0.012} \\
& LPIPS~\cite{zhang2018unreasonable}~($\downarrow$) 
& 0.169\pmgray{0.057}
& \cellcolor{myblue}0.251\pmgray{0.049}
& 0.358\pmgray{0.086}
& 0.487\pmgray{0.076}
& 0.678\pmgray{0.246}
& \cellcolor{myblue}0.251\pmgray{0.052}
& \cellcolor{mygreen}\textbf{0.243}\pmgray{0.085} \\
& ArcFace~\cite{deng2019arcface}~($\uparrow$)
& 0.943\pmgray{0.037}
& \cellcolor{myblue}0.564\pmgray{0.144}
& 0.238\pmgray{0.144}
& 0.540\pmgray{0.061}
& 0.193\pmgray{0.184}
& 0.358\pmgray{0.123}
& \cellcolor{mygreen}\textbf{0.566}\pmgray{0.210} \\
\midrule

\multirow{4}{*}{Big Nose}
& EPR~($\uparrow$) 
& 0.207\pmgray{0.280}
& 0.882\pmgray{0.539}
& 0.938\pmgray{0.608}
& \cellcolor{myblue}1.035\pmgray{0.775}
& 0.793\pmgray{0.600}
& 0.867\pmgray{0.627}
& \cellcolor{mygreen}\textbf{1.060}\pmgray{0.608} \\
& CLIP-Score~($\uparrow$)
& 0.140\pmgray{0.019}
& 0.148\pmgray{0.014}
& \cellcolor{myblue}0.173\pmgray{0.017}
& 0.168\pmgray{0.012}
& \cellcolor{mygreen}\textbf{0.200}\pmgray{0.029}
& 0.157\pmgray{0.014}
& 0.159\pmgray{0.012} \\
& LPIPS~($\downarrow$)
& 0.169\pmgray{0.057}
& 0.270\pmgray{0.055}
& 0.275\pmgray{0.082}
& 0.546\pmgray{0.078}
& 0.699\pmgray{0.231}
& \cellcolor{myblue}0.208\pmgray{0.048}
& \cellcolor{mygreen}\textbf{0.191}\pmgray{0.050} \\
& ArcFace~($\uparrow$)
& 0.943\pmgray{0.037}
& \cellcolor{myblue}0.545\pmgray{0.112}
& 0.463\pmgray{0.182}
& 0.487\pmgray{0.084}
& 0.200\pmgray{0.191}
& 0.412\pmgray{0.121}
& \cellcolor{mygreen}\textbf{0.761}\pmgray{0.059} \\
\midrule

\multirow{4}{*}{Young}
& EPR~($\uparrow$)
& 0.773\pmgray{0.619}
& 1.411\pmgray{0.909}
& \cellcolor{myblue}1.685\pmgray{1.220}
& 0.937\pmgray{0.770}
& 1.427\pmgray{0.709}
& 1.536\pmgray{0.989}
& \cellcolor{mygreen}\textbf{1.817}\pmgray{1.117} \\
& CLIP-Score~($\uparrow$)
& 0.113\pmgray{0.013}
& 0.180\pmgray{0.014}
& \cellcolor{mygreen}\textbf{0.211}\pmgray{0.014}
& 0.184\pmgray{0.013}
& 0.204\pmgray{0.017}
& 0.184\pmgray{0.013}
& \cellcolor{myblue}0.206\pmgray{0.013} \\
& LPIPS~($\downarrow$)
& 0.169\pmgray{0.057}
& \cellcolor{mygreen}\textbf{0.258}\pmgray{0.065}
& 0.343\pmgray{0.082}
& 0.567\pmgray{0.082}
& 0.678\pmgray{0.246}
& 0.309\pmgray{0.052}
& \cellcolor{myblue}0.305\pmgray{0.082} \\
& ArcFace~($\uparrow$)
& 0.943\pmgray{0.037}
& \cellcolor{myblue}0.546\pmgray{0.139}
& 0.306\pmgray{0.139}
& \cellcolor{mygreen}\textbf{0.646}\pmgray{0.095}
& 0.266\pmgray{0.191}
& 0.357\pmgray{0.128}
& 0.387\pmgray{0.142} \\
\midrule

\multirow{4}{*}{Beards}
& EPR~($\uparrow$)
& 0.763\pmgray{0.669}
& 1.165\pmgray{0.974}
& 1.739\pmgray{1.268}
& 0.919\pmgray{0.383}
& \cellcolor{myblue}2.059\pmgray{1.541}
& 1.255\pmgray{0.884}
& \cellcolor{mygreen}\textbf{2.161}\pmgray{1.080} \\
& CLIP-Score~($\uparrow$)
& 0.112\pmgray{0.033}
& 0.134\pmgray{0.038}
& \cellcolor{myblue}0.206\pmgray{0.047}
& 0.132\pmgray{0.032}
& 0.204\pmgray{0.022}
& 0.141\pmgray{0.031}
& \cellcolor{mygreen}\textbf{0.211}\pmgray{0.043} \\
& LPIPS~($\downarrow$)
& 0.169\pmgray{0.057}
& 0.274\pmgray{0.054}
& 0.353\pmgray{0.090}
& 0.490\pmgray{0.073}
& 0.741\pmgray{0.216}
& \cellcolor{myblue}0.257\pmgray{0.053}
& \cellcolor{mygreen}\textbf{0.233}\pmgray{0.071} \\
& ArcFace~($\uparrow$)
& 0.943\pmgray{0.037}
& \cellcolor{myblue}0.550\pmgray{0.116}
& 0.294\pmgray{0.164}
& 0.542\pmgray{0.058}
& 0.171\pmgray{0.191}
& 0.422\pmgray{0.136}
& \cellcolor{mygreen}\textbf{0.562}\pmgray{0.260} \\
\midrule

\multirow{4}{*}{Blond Hair}
& EPR~($\uparrow$)
& 0.110\pmgray{0.032}
& 0.876\pmgray{0.794}
& \cellcolor{myblue}1.173\pmgray{1.002}
& 0.760\pmgray{0.548}
& \cellcolor{myblue}1.173\pmgray{0.678}
& 1.302\pmgray{1.014}
& \cellcolor{mygreen}\textbf{1.386}\pmgray{1.213} \\
& CLIP-Score~($\uparrow$)
& 0.143\pmgray{0.029}
& 0.153\pmgray{0.030}
& \cellcolor{mygreen}\textbf{0.186}\pmgray{0.039}
& 0.143\pmgray{0.027}
& \cellcolor{myblue}0.185\pmgray{0.021}
& 0.154\pmgray{0.033}
& 0.170\pmgray{0.036} \\
& LPIPS~($\downarrow$)
& 0.169\pmgray{0.057}
& 0.262\pmgray{0.052}
& 0.303\pmgray{0.082}
& 0.491\pmgray{0.076}
& 0.713\pmgray{0.208}
& \cellcolor{myblue}0.227\pmgray{0.103}
& \cellcolor{mygreen}\textbf{0.221}\pmgray{0.072} \\
& ArcFace~($\uparrow$)
& 0.943\pmgray{0.037}
& \cellcolor{myblue}0.547\pmgray{0.117}
& 0.409\pmgray{0.210}
& 0.544\pmgray{0.053}
& 0.169\pmgray{0.128}
& 0.387\pmgray{0.141}
& \cellcolor{mygreen}\textbf{0.686}\pmgray{0.155} \\


\bottomrule
\end{tabular}
}
\caption{Our explainable editing framework achieves superior attribute manipulation performance without degrading perceptual quality or identity consistency. We evaluate five facial attributes using EPR, CLIP-Score, LPIPS, and ArcFace across all methods. We highlight the best value in \colorbox{mygreen}{\textbf{green}} \textbf{with bold text} and the second best value in \colorbox{myblue}{blue}.
}
\label{tab:statistic_result}
\end{table*}

\subsection{Editing Precision Ratio (EPR)}
\label{sec:metric}

A desirable causal probing metric should induce significant changes in the target attribute while minimizing unintended modifications to other (non-target) attributes. However, existing evaluation metrics do not jointly measure the degree of the \emph{editing effectiveness} on the target attribute and the side effects on the non-target attributes. 
To address this, we propose a metric called \emph{Editing Precision Ratio} (EPR), which provides a unified and interpretable metric.

Given $N$ pairs of original and edited images $\{x_i, x_i'\}_{i=1}^N$, let $f_{\mathrm{target}}(\cdot)$ denote the classifier score for the target attribute, and $f_j(\cdot)$ denote the classification score for the $j$-th non-target attribute (for $j = 1,\ldots,L$).

\paragraph{Average Target Attribute Change (Editing Effectiveness).} This term captures how much the editing on the target attribute shifts the score of the target attribute:
\begin{equation}
\label{eq:delta_target}
  \Delta_{\mathrm{target}} = \frac{1}{N}\sum_{i=1}^N \left| f_{\mathrm{target}}(x_i') - f_{\mathrm{target}}(x_i) \right|
\end{equation}

\paragraph{Average Non-Target Attribute Change (Side Effects).} This term reflects the average change across all unrelated attributes, measuring how much collateral change the edit induces on all the non-target attributes:
\begin{equation}
\label{eq:delta_non_target}
  \Delta_{\mathrm{non\_target}} = \frac{1}{L} \sum_{j=1}^L \left( \frac{1}{N}\sum_{i=1}^N \left| f_j(x_i') - f_j(x_i) \right| \right)
\end{equation}

\paragraph{Editing Precision Ratio.} The final score is defined as the ratio between the average target attribute change and the average non-target attribute change:
\begin{equation}
  \mathrm{EPR} = \frac{\Delta_{\mathrm{target}}}{\Delta_{\mathrm{non\_target}} + \epsilon}
\end{equation}
where \( \epsilon \) is a small constant added for numerical stability. 

A higher EPR indicates that the editing method achieves greater changes in the target attribute with less unintended change to non-target attributes, thus providing a quantitative measure of both editing effectiveness on the target attribute and side effects on the non target attributes. In practice, $f_j$ denotes the classifier logits rather than probabilities, as logits preserve relative semantic strength without saturation. We set $\epsilon = 10^{-8}$ for numerical stability.

\section{Experiments}
\label{sec:exps}

\subsection{Implementation Details}

We evaluate CASL on multiple datasets, including FFHQ~\cite{karras2019style}, CelebA-HQ~\cite{karras2017progressive}, LSUN-Church~\cite{yu2015lsun} and AFHQ~\cite{choi2020stargan}. For each dataset, we adopt the corresponding diffusion backbone: DDPM++~\cite{song2020score} for FFHQ and CelebA-HQ, DDPM~\cite{ho2020denoising} for LSUN and iDDPM~\cite{nichol2021improved} for AFHQ. All diffusion backbones are official pretrained checkpoints and remain frozen. 
We uniformly sample 50 timesteps between \(t = 999\) and \(t = 500\) and extract the associated U-Net activations. The Sparse Autoencoder is trained on bottleneck activations, after which we learn a lightweight linear mapping for each concept using 1{,}000 real images. Full training configurations and runtime statistics are provided in the Appendix.
All experiments are run on 1 NVIDIA RTX A100 GPU (80GB memory). More implementation details can be found in Appendix.

\subsection{Visual Validation of Concept Alignment}
\label{sec:effect}

Figure~\ref{fig:main_result} provides a qualitative validation of whether the
concept-aligned latent directions learned by CASL correspond to the intended
semantics and whether the resulting interventions remain localized and
disentangled.

Across CelebA-HQ, the aligned directions consistently induce the intended
semantic change (e.g., \emph{Smiling}, \emph{Young}, \emph{Beards}) while leaving
identity, background, and pose largely unchanged.  
This indicates that the learned latent units capture concept-specific information
rather than mixed or correlated factors.

On AFHQ-Dog and LSUN-Church, CASL-Steer exhibits a similar behavior:
concept directions such as \emph{Smiling Dog}, \emph{Puppy}, \emph{Gothic}, and
\emph{Wooden} produce coherent semantic shifts while preserving viewpoint,
geometry, and low-level appearance.  
The stability of the edits across categories suggests that the aligned latent
space carries transferable and structurally meaningful concept directions.

Overall, these observations validate that the aligned latent dimensions learned by CASL correspond to stable and interpretable semantic factors.  
Manipulating a single dimension produces a targeted shift along the intended
concept with little influence on unrelated factors, indicating that CASL
successfully separates concept-relevant information from background variation.

Additional examples are provided in the Appendix to further confirm that
CASL captures clean and interpretable concept directions with minimal
collateral change.

\subsection{Quantitative Evaluation}
Our goal is not to introduce a new editing algorithm, but to use semantic editing as a \textbf{causal probe} for verifying whether a supervised sparse latent direction truly corresponds to a human-defined concept. We therefore compare CASL-Steer against representative attribute-editing methods that operate through activation-space or directional manipulations, including BoundaryDiffusion~\cite{zhu2023boundary}, Asyrp~\cite{kwon2022diffusion}, Concept Slider~\cite{2024concept}, MasaCtrl~\cite{cao2023masactrl}, and SwiftEdit~\cite{Nguyen_2025_CVPR}. We report four metrics commonly used in facial attribute editing: CLIP-Score~\cite{hessel2021clipscore} for semantic alignment, LPIPS~\cite{zhang2018unreasonable} for perceptual distortion, ArcFace similarity~\cite{deng2019arcface} for identity preservation, and our proposed EPR (Sec.~\ref{sec:metric}) for measuring the trade-off between attribute strength and unintended changes.


Table~\ref{tab:statistic_result} shows results on CelebA-HQ. 
For each attribute, we randomly sample 32 images from the test split and use only the top-1 aligned latent dimension from CASL. 
CASL-Steer injects the concept-aligned shift once during the DDIM sampling process, while all baseline methods follow their standard configurations. 
Additional results with more datasets and a broader range of concepts are provided in the Appendix.

\vspace{0.5em}
\noindent\textbf{Reconstruction Baseline.}
The ``Recon.'' baseline corresponds to passing the input image through the inversion/reconstruction pipeline without applying any concept direction. This baseline is \emph{not} an identity mapping, as the DDIM inversion and U-Net bottleneck reconstruction introduce small but non-negligible deviations in $h$-space. Consequently, its EPR is non-zero. We include this baseline to contextualize how much attribute drift arises purely from the inversion process, separating it from concept-specific changes.

\vspace{0.5em}
\noindent\textbf{Editing Strength.}
Across all concepts, CASL-Steer achieves competitive or superior CLIP-Score compared to LoRA-based (Slider) and training-free (SwiftEdit) methods, despite not updating the diffusion model. This indicates that the concept-aligned sparse directions learned by CASL capture the target semantic effect reliably.

\vspace{0.5em}
\noindent\textbf{Unintended Changes.}
CASL-Steer exhibits markedly fewer unintended modifications, as reflected by substantially lower LPIPS and higher ArcFace similarity. This suggests that aligning sparse latent dimensions with human-defined concepts leads to more localized and disentangled effects, while black-box methods often introduce changes correlated with the target concept (e.g., identity shift for ``smiling'' or background leakage for ``blond hair'').

\vspace{0.5em}
\noindent\textbf{Editing Precision (EPR).}
Our EPR metric jointly measures target-attribute strength and collateral distortion. CASL-Steer consistently achieves the highest EPR across all concepts, demonstrating that concept-aligned sparse latents enable more precise and cleaner semantic shifts. Notably, CASL-Steer achieves the largest improvements on attributes that are typically entangled with other facial factors (e.g., \emph{young} often correlates with smooth skin, and \emph{blond hair} correlates with specific color tones). This suggests that supervised concept alignment helps separate the target attribute from its co-occurring correlated features.

\subsection{Evaluation of Sparse Feature Representations}
\label{sec:sae_eval}
We evaluate the reconstruction quality and sparsity of the sparse autoencoder (SAE) representations as indicators of interpretability. Reconstruction is measured by mean squared error (MSE) and cosine similarity between original and reconstructed activations. Sparsity is quantified by the dimension activation ratio (DAR):

\begin{equation}
\mathrm{DAR}(z) = \frac{1}{K} \sum_{j=1}^K \mathbb{I}\left[ \frac{1}{N} \sum_{i=1}^N z_{ij} > \tau \right]
\end{equation}
where $z \in \mathbb{R}^{N \times K}$ is the latent activation matrix and $\tau$ is a fixed threshold. Here, $K$ is determined by a predefined expansion ratio $\gamma_{\text{sae}}$ multiplied by the input channel dimension $C$. Lower DAR indicates a sparser representation.


\begin{table}[ht]
\centering
\footnotesize
\resizebox{\columnwidth}{!}{
\begin{tabular}{lccc}
\toprule
\textbf{Params} 
& \textbf{MSE($\times 10^{-2}$)} $\downarrow$ 
& \textbf{Cosine Sim.} $\uparrow$ 
& \textbf{DAR ($\times 10^{-2}$)} $\downarrow$ \\
\midrule
$\gamma_{\text{sae}}=16$  & 9.27  & 0.965 & 6.32  \\
$\gamma_{\text{sae}}=32$  & 5.39  & 0.980 & 3.54  \\
$\gamma_{\text{sae}}=64$  & 3.19  & 0.988 & 1.87  \\
$\gamma_{\text{sae}}=128$ & 1.91  & 0.993 & 1.08  \\
$\gamma_{\text{sae}}=256$ & 0.94  & 0.996 & 0.70  \\
$\gamma_{\text{sae}}=512$ & 0.56  & 0.998 & 0.39  \\
\bottomrule
\end{tabular}
}
\caption{
Reconstruction quality and sparsity of the SAE across different expansion ratios
$\gamma_{\text{sae}}$. 
MSE and DAR values are scaled by $10^{-2}$ for readability.
}

\label{tab:sae_eval}
\end{table}

\begin{figure*}[ht]
  \centering
  \begin{subfigure}[t]{0.245\linewidth}
    \centering
    \includegraphics[width=\linewidth]{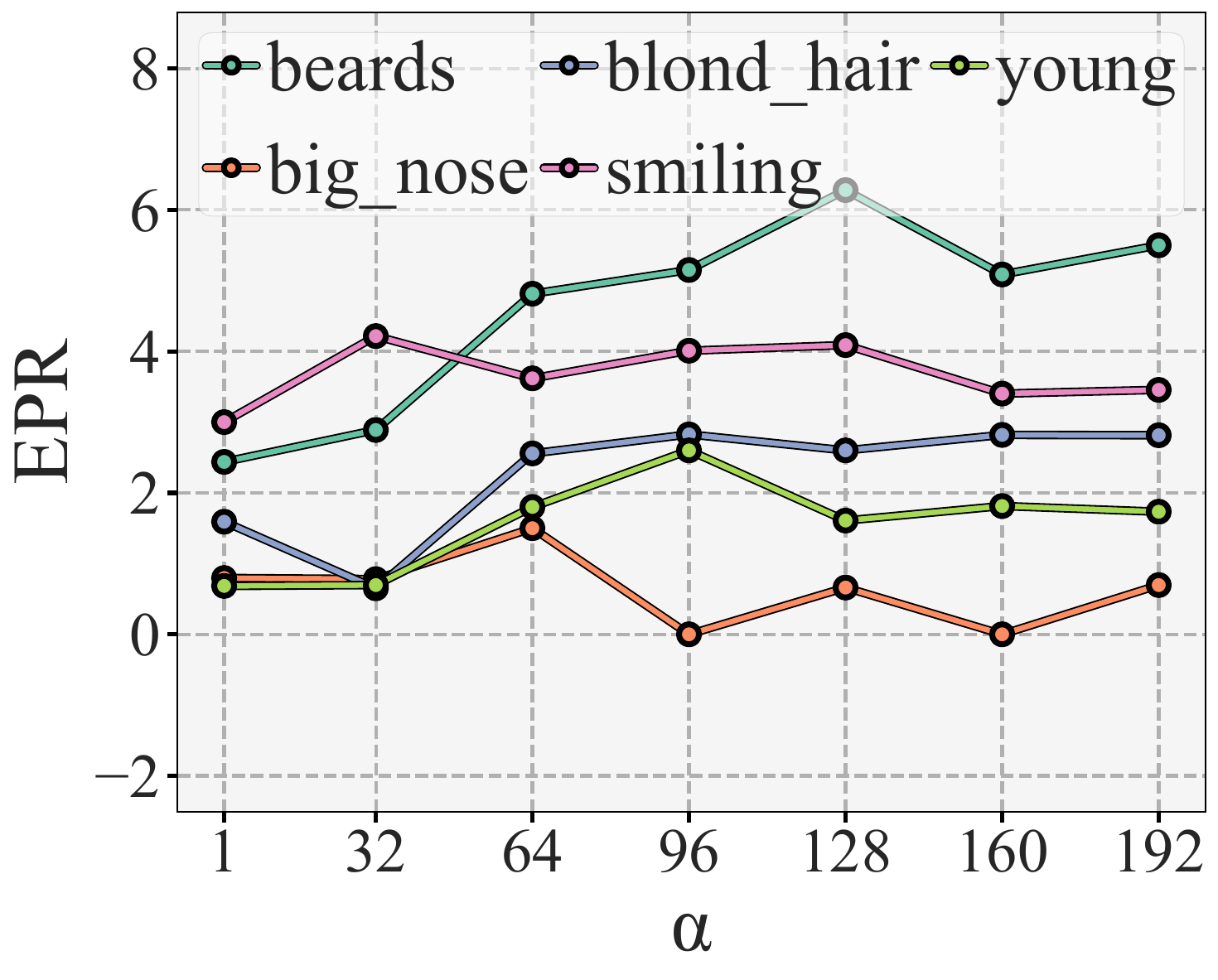}
    \caption{EPR vs $\alpha$ (Concepts)}
    \label{fig:hyper_a}
  \end{subfigure}
  \hfill
  \begin{subfigure}[t]{0.245\linewidth}
    \centering
    \includegraphics[width=\linewidth]{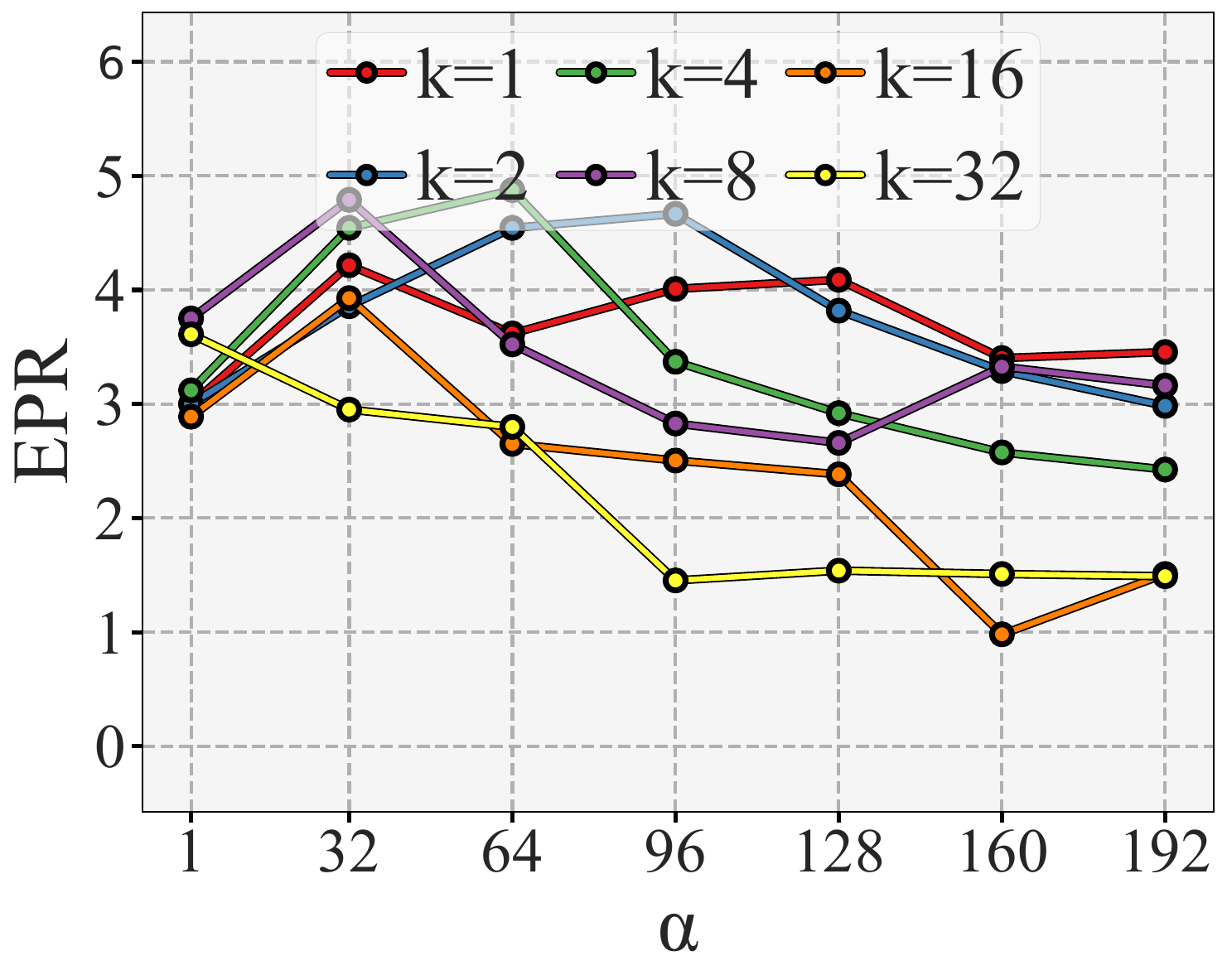}
    \caption{EPR vs $\alpha$ (Top-$k$)}
    \label{fig:hyper_b}
  \end{subfigure}
  \hfill
  \begin{subfigure}[t]{0.245\linewidth}
    \centering
    \includegraphics[width=\linewidth]{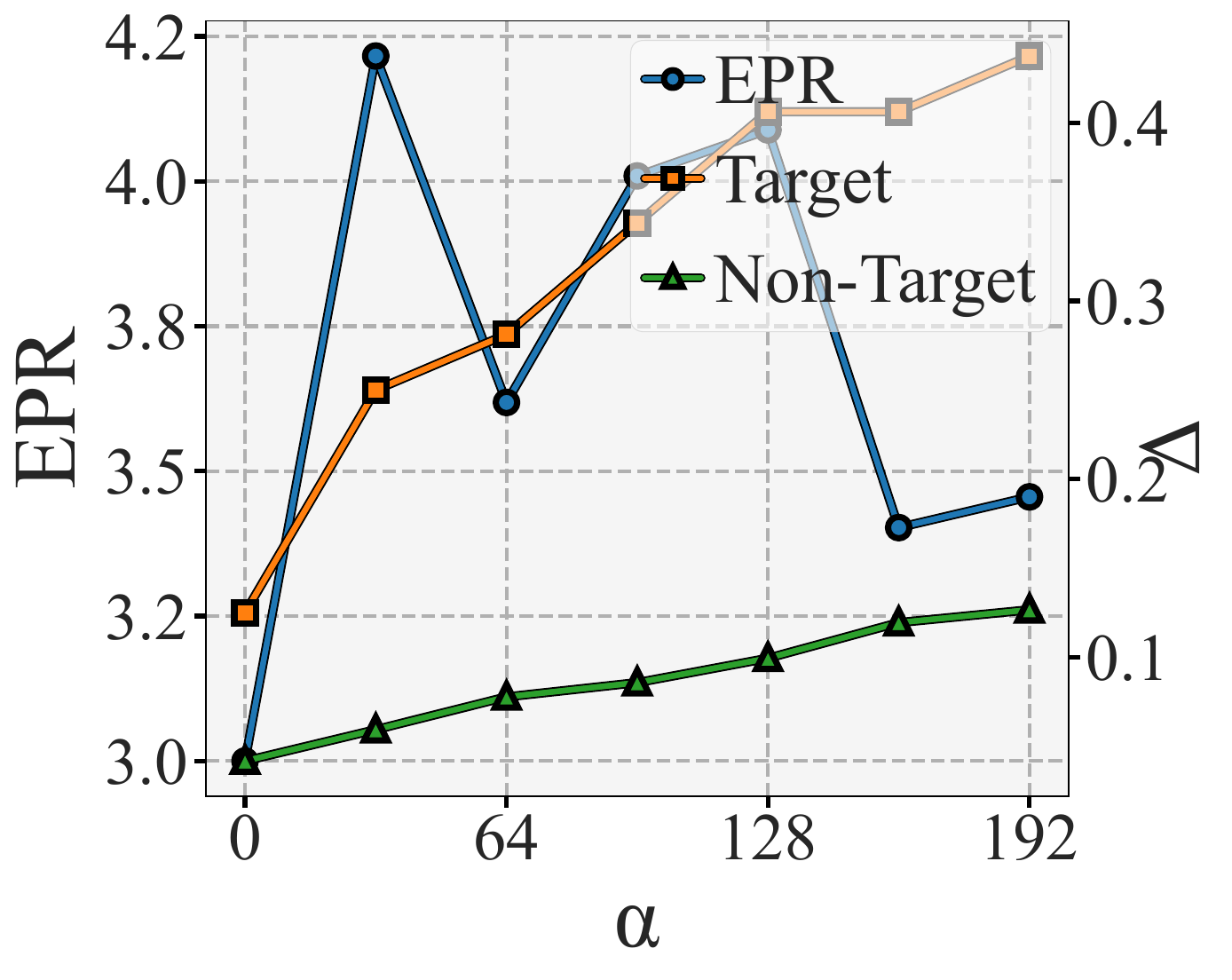}
    \caption{EPR on smiling (Top-$k{=}1$)}
    \label{fig:hyper_c}
  \end{subfigure}
  \hfill
  \begin{subfigure}[t]{0.245\linewidth}
    \centering
    \includegraphics[width=\linewidth]{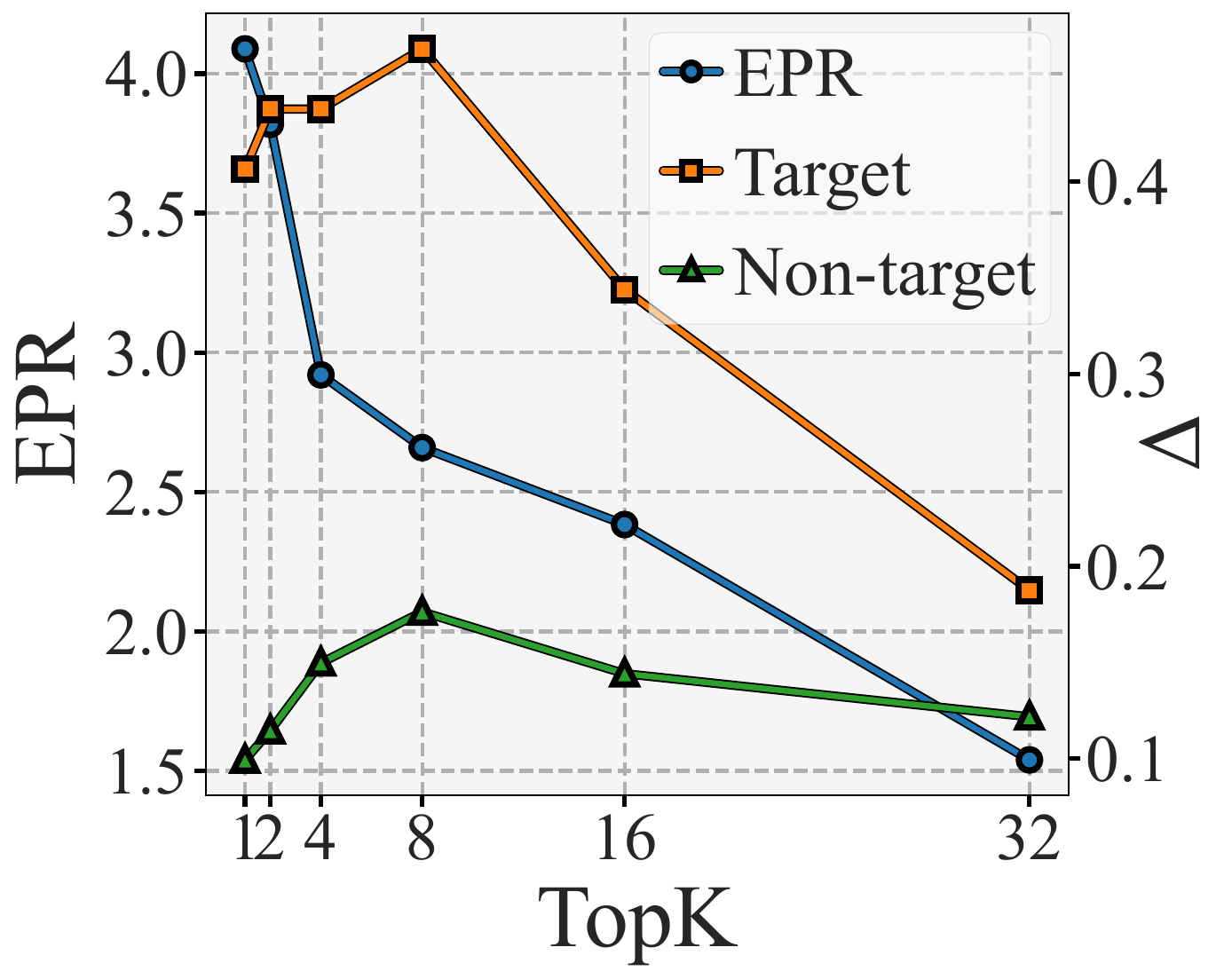}
    \caption{EPR on smiling ($\alpha{=}128$)}
    \label{fig:hyper_d}
  \end{subfigure}
  \caption{
    Hyperparameter analysis of editing intensity $\alpha$ and top-$k$ dimension selection. (a) Single-dimension editing shows stable EPR across concepts. (b) Multi-dimension editing reduces precision with increasing $\alpha$. (c) Target and non-target changes scale proportionally for $k=1$. (d) Increasing $k$ degrades EPR, confirming sparse editing maximizes interpretability.
  }
  \label{fig:hyper_analysis}
\end{figure*}

Table~\ref{tab:sae_eval} reports SAE performance across different expansion ratios $\gamma_{\text{sae}}$ on CelebA-HQ, with input activations $h^{(t)}$ sampled uniformly from 50 timesteps between $t=999$ and $t=500$ during the DDIM inversion process. All configurations achieve low reconstruction error while maintaining strong sparsity, demonstrating the effectiveness of our design. Implementation details of the SAE and additional experiments with varying $\lambda_{\text{sparse}}$ values are provided in the Appendix. We adopt $\gamma_{\text{sae}}=128$ and sparsity weight $\lambda_{\text{sparse}}=32$ as default for balancing performance and efficiency.

\subsection{Concept Alignment Evaluation}
\label{sec:alignment}


To evaluate whether CASL successfully aligns specific latent dimensions $z_i$ to human-defined concepts, we perform a classification-based probing experiment. For each concept $c$, we select the top-$k$ latent units according to $\mathcal{I}_c$ (Eq.~\ref{eq:index_selection}) and train a linear SVM classifier using these $k$-dimensional features.

As is shown in Table~\ref{tab:alignment_result}, the experiments are conducted under the hyperparameter setting of $\gamma_{\text{sae}}=128$ and $\lambda_{\text{sparse}}=32$, resulting in a high-dimensional latent space of $512 \times 128 = 65,536$ dimensions. The SVM classifiers are trained and tested on balanced datasets containing 1,000 positive and 1,000 negative samples each. During the inversion process, activation values $h^{(t)}$ are uniformly sampled at 50 time steps between $t=999$ and $t=500$, which serve as inputs to the classifiers. The dataset is split into training and testing sets with an 8:2 ratio.

\begin{table}[ht]\footnotesize
\centering
\begin{tabular}{l *{6}{c}}
\toprule
\textbf{Concepts}&\textbf{top-1}&\textbf{top-2}&\textbf{top-4}&\textbf{top-8}&\textbf{top-16}

\\
\midrule
Smiling& 0.671&0.755&0.851&0.948&0.993\\
Big Nose&0.675&0.694&0.736&0.771&0.818\\
Young &0.655&0.702&0.757&0.797&0.855\\
No Beard&0.757&0.825&0.851& 0.891&0.929\\
Blond Hair&0.677&0.750&0.843&0.883&0.938\\
\bottomrule
\end{tabular}
\caption{SVM classification accuracy for top-$k$ latent dimensions across different concepts.}
\label{tab:alignment_result}
\end{table}

\noindent\textbf{Results.}
Even the single highest-ranked latent unit (top-1) achieves accuracy vastly above the random baseline ($50\%$), confirming that individual CASL-aligned dimensions carry strong concept-specific signals. Increasing $k$ further improves accuracy, indicating that additional aligned units contribute complementary semantic cues. Using only 16 latent units per concept, all classifiers approach near-perfect accuracy ($>0.93$), demonstrating that the aligned latent space is highly concentrated and semantically coherent. These results validate that CASL effectively isolates concept-relevant latent factors in a compact subset of sparse units.


\subsection{Hyperparameter Analysis}
\label{sec:hyper}

To better understand the behavior and robustness of our method, we conduct ablation studies of two key hyperparameters: the number of top-$k$ latent dimensions for editing (Eq.~\ref{eq:index_selection}), and the editing intensity coefficient $\alpha$ (Eq.~\ref{eq:s3_main}). 
  

Our results are shown in Fig~\ref{fig:hyper_analysis}.\textbf{(a)}. When only the top-1 latent dimension is edited (Fig.~\ref{fig:hyper_a}), EPR remains stable across a wide range of $\alpha$ for most concepts. This indicates that our SAE successfully extracts disentangled and interpretable latent factors, amplifying a single direction increases the target attribute without introducing unrelated concepts, even under strong editing. 
\textbf{(b)}~For the \textit{smiling} attribute (Fig.~\ref{fig:hyper_b}), we observe that EPR is robust to changes in $\alpha$ only for top-$k$=1. When $k>1$, EPR decreases significantly as $\alpha$ increases, suggesting that editing multiple latent dimensions simultaneously inevitably entangles additional semantic information and reduces editing precision. \textbf{(c)}~Figure~\ref{fig:hyper_c} further decomposes the editing behavior for \textit{smiling} ($k=1$). As $\alpha$ increases, both the $\Delta_{\mathrm{target}}$ (Eq.~\ref{eq:delta_target}) and $\Delta_{\mathrm{non\_target}}$ (Eq.~\ref{eq:delta_non_target}) changes rise, but EPR remains nearly constant. This implies that, although overall attribute changes are magnified by stronger editing, the \textit{relative} specificity of the edit (i.e., the ratio between target and side effect) is well preserved due to the high semantic purity of the selected latent dimensions.
\textbf{(d)}~Finally, Figure~\ref{fig:hyper_d} explores how EPR and score changes respond to increasing $k$ at a fixed $\alpha$. As $k$ increases, EPR steadily declines and the gap between target and non-target changes narrows. This supports our design choice that sparsity in editing directions (small $k$) is crucial for maximizing interpretability and precision in semantic manipulation.

Overall, our method yields purified and accurately aligned latent features, allowing precise semantic editing with minimal change. Editing more latent dimensions, however, introduces entanglement and diminishes editing effectiveness. More results can be found in Appendix.

\subsection{Computational Cost}
CASL requires training the sparse autoencoder once on U-Net activations. 
On a single A100 GPU, SAE training takes approximately 6 hours with a memory footprint below 32 GB. 
The subsequent concept-mapping step is a lightweight linear regression and completes within minutes. 
CASL-Steer itself adds negligible cost at inference, since it injects a single latent shift during DDIM sampling. 
Overall, CASL introduces only a one-time offline training cost, and its per-edit overhead is comparable to standard DDIM generation.






%

\section{Conclusion}
\label{sec:conclusion}

We introduced CASL, a framework that learns a sparse latent space whose individual dimensions are aligned with human-defined semantic concepts in diffusion models. The resulting representation makes concept-relevant information more explicit and separates it from other latent factors, providing a clearer view of how semantics are organized inside the model. Building on this space, CASL-Steer enables controlled probing of concept influence through simple latent interventions, offering a direct way to examine how specific attributes affect the generative process without modifying the diffusion model itself. We further proposed EPR, a metric that jointly evaluates the intended semantic change and the unintended shifts in non-target attributes, allowing a more balanced assessment of semantic alignment quality. Our experiments show that CASL yields compact and interpretable concept directions, and that CASL-Steer consistently exposes their semantic effects across datasets. We hope these insights encourage further exploration of structured and interpretable latent representations for diffusion models.
{
    \small
    \bibliographystyle{ieeenat_fullname}
    \bibliography{main}
}

\clearpage
\setcounter{page}{1}
\maketitlesupplementary

\section{Implementation Details}
\subsection{Code}
We provide our code in the supplementary material. Please check.

\subsection{Model and Architecture Settings}

\paragraph{SAE Training Details.}
We train the sparse autoencoder (SAE) on cached U\hspace{0.3mm}-Net bottleneck 
activations extracted at a fixed timestep $t_0$. 
Each activation $h \in \mathbb{R}^{(H W) \times C}$ is reshaped into 
$N = H W$ spatial tokens of dimension $C$. 
Following prior work on token-level interpretability, we adopt a 
\emph{token-wise} training scheme: a minibatch of 64 images yields 
$64 \times N$ training tokens per optimization step, allowing the SAE 
to fully leverage spatial diversity and substantially increase the number of 
effective training samples.

We keep both the sparsity weight $\alpha$ fixed during 
training; no annealing schedule is applied. We did not observe training 
instability that would require warm-up or staged sparsification.

\textbf{Architecture.}
The SAE is a linear encoder--decoder with learnable pre-bias and latent bias:
\[
z = \phi\!\left(W_{\eta}(h - b_{\text{pre}}) + b_{z}\right), 
\qquad
\hat{h} = W_{\psi} z + b_{\text{pre}},
\]
where $W_{\eta} \in \mathbb{R}^{K \times C}$ and $W_{\psi} \in \mathbb{R}^{C \times K}$.
We use \emph{untied} weights ($W_{\psi} \neq W_{\eta}^{\!\top}$), as tying the decoder noticeably 
reduces reconstruction fidelity at large expansion ratios.
The latent dimension is set to $K = C \cdot \text{scale}$ with $\text{scale}=32$. 
We adopt \texttt{ReLU} in our main experiments due to its training stability 
and smoother gradients.

\textbf{Timestep Embedding.}
We keep the timestep embedding enabled 
(\texttt{ignore\_timesteps = False}) so that the SAE can capture mild temporal variation 
in U\hspace{0.2mm}-Net activations between denoising timesteps $t \in [500,999]$. 
Although this temporal variation is limited, we find that including timestep information slightly improves reconstruction accuracy without affecting sparsity or alignment.

\textbf{Objective.}
The training objective follows:
\[
\mathcal{L}
    = \bigl\| \hat{h} - h \bigr\|_2^2
    + \lambda_{\text{sparse}} \cdot \| z \|_{1},
\]
where the sparsity loss is implemented as the mean absolute activation of $z$.
We tune $\lambda_{\text{sparse}}$ to balance reconstruction fidelity and latent sparsity.

To validate the generalizability of our sparse autoencoder, we further conduct ablation studies on the \textbf{FFHQ dataset}. As shown in Table~\ref{tab:sae_eval}, we report the reconstruction quality (MSE, Cosine Similarity) and sparsity (DAR, $\tau=0.01$) under different sparsity loss weights $\lambda_{\text{sparse}}$, with the latent dimension $\gamma_{\text{sae}}$ fixed at $128$. Compared to the results on CelebA-HQ (see main text), the trends on FFHQ are consistent, demonstrating the robustness of our method across datasets.

\textbf{Optimization.}
The SAE is trained for 100 epochs using Adam 
(learning rate $\texttt{lr\_sae} = 5\times10^{-4}$, weight decay disabled). 
We use a batch size of 64 images (expanded to 
$64 \times H W$ latent tokens per step). 
Gradient updates are performed on all tokens jointly, and 
model checkpoints are saved every 10 epochs. 
This token-wise design enables efficient large-scale SAE training 
while maintaining high reconstruction accuracy and strong sparsity.


\subsection{Stage-2: Concept Alignment MLP}
To map the sparse latent representation to the activation direction for semantic editing, we use a simple linear transformation implemented as a single-layer perceptron:
\[
\Delta h = W_\Delta z + b_\Delta
\]
where $z$ is the latent vector, $W_\Delta$ and $b_\Delta$ are learnable parameters, and $\Delta h$ is the predicted direction in the activation space.

This operation is implemented in PyTorch as:
\begin{lstlisting}[language=Python]
import torch.nn as nn

direction_decoder = nn.Linear(latent_dim, out_dim)
delta_h = direction_decoder(z)
\end{lstlisting}
\subsection{Steerding Pipeline.}
We apply semantic intervention during the denoising trajectory of the
U\hspace{0.15mm}-Net.  
Let $t_{\mathrm{edit}} = 500$ denote the editing threshold.  
Following DDIM inversion, we obtain the noisy trajectory 
$\{x_t\}_{t=999}^{0}$ and at each denoising step we extract the bottleneck
activation $h_t$ from the middle block.

\textbf{Timestep-wise intervention.}
Editing is applied not at a single timestep, but repeatedly across the range
$t \in [500,999]$.  
At every denoising step satisfying $t \ge t_{\mathrm{edit}}$, we modify the
bottleneck activation through a concept-aligned shift:
\[
h_t' = h_t + \gamma\,\Delta h_c, \qquad 
\Delta h_c = W_{\Delta}\,z_t,
\]
where $z_t$ is the sparse latent representation produced by the SAE encoder,
and $W_{\Delta}$ is the concept-mapping linear layer from Stage~2.
Only this middle-block activation is modified; all other layers and all
timesteps $t < 500$ remain unchanged.

\textbf{Continuation of sampling.}
The modified activation $h_t'$ is injected via a forward hook and the remaining
DDIM update is computed normally:
\[
x_{t-1}
= \sqrt{\alpha_{t-1}}\,P_t\!\left(\epsilon_\theta(x_t \mid h_t')\right)
 + D_t\!\left(\epsilon_\theta(x_t)\right).
\]
Since the intervention is applied across roughly 50 denoising steps (using a
step size of 10 from $t=999$ to $t=500$), the influence of the concept-aligned
shift accumulates smoothly along the trajectory, enabling controlled semantic
modification without altering the early generative dynamics.

\subsection{Layer Selection for \emph{h}-space in U-Net}

Following prior work~\cite{kwon2022diffusion}, we define the \emph{h}-space as the intermediate feature representation at the bottleneck layer of the U-Net architecture. Specifically, we select the 8th layer (with spatial resolution $8 \times 8$ and channel dimension 512) as our \emph{h}-space. This layer is not directly influenced by skip connections, which ensures that the activations encode high-level and abstract semantics, making it suitable for both sparse autoencoding and concept editing. 

Our choice is motivated by the comprehensive empirical analysis in~\cite{kwon2022diffusion}, where the authors systematically evaluated semantic editing performance at different layers of the U-Net. Their experiments demonstrated that the bottleneck layer (layer 8) contains the richest and most disentangled semantic information, and editing this layer yields the most effective and interpretable results for semantic manipulation tasks.

\subsection{Hyperparameter Selection for Training Loss.}
The selection of hyperparameters for CLIP loss (Eq.~\ref{eq:cliploss}) is crucial for achieving optimal semantic alignment and faithful reconstruction in semantic editing tasks. As shown in Table~\ref{tab:clip_hyper}, we empirically select the weighting factors for the CLIP loss and reconstruction loss ($\lambda_{\text{CLIP}} : \lambda_{\text{recons}}$) based on experiments and prior work, tailoring the settings to different datasets and target concepts. For each dataset (e.g., CelebA-HQ, FFHQ, LSUN-Church, AFHQ-Dog), we consider multiple representative concepts and adjust the loss weights accordingly to balance the semantic editing strength and image fidelity. Table~\ref{tab:clip_hyper} summarizes the datasets, concepts, target descriptions, and corresponding hyperparameter configurations used in our experiments.

\subsection{Datasets and Preprocessing}
\section*{Datasets and Preprocessing}

\paragraph{AFHQ.}
We use the AFHQ–Dog subset following prior diffusion editing literature.
Images are loaded from the official train/test split
(\texttt{train/dog/*.png}, \texttt{test/dog/*.png}).  
Each image is resized to $s\times s$ (where $s$ is the target resolution for each
backbone, e.g., $256$ or $512$), converted to a float tensor in $[0,1]$, and
normalized to $[-1,1]$ using a standard
$\texttt{ToTensor} + \texttt{Normalize((0.5,0.5,0.5),(0.5,0.5,0.5))}$ transform.
No data augmentation is applied to preserve the deterministic nature of DDIM inversion.

\paragraph{CelebA–HQ.}
CelebA–HQ is stored in LMDB format for efficient random access.
Each sample is retrieved via a resolution-specific LMDB key 
\texttt{``<res>-<index>''}, decoded into a JPEG/PNG image, and processed with the 
same normalization described above.  
The LMDB files contain the official train and test partitions.  
All images are resized to the target model resolution and transformed with
$\texttt{ToTensor}$ and $\texttt{Normalize}$ without augmentation.

\paragraph{LSUN-Church.}
For LSUN, we adopt the official LMDB release and follow the conventional 
train/validation split (\texttt{church\_outdoor\_train}, \texttt{church\_outdoor\_val}).
The loader retrieves each image from LMDB, converts it to RGB, applies
$\texttt{Resize}$ and $\texttt{CenterCrop}$ to match the desired resolution, and then
normalizes it to the $[-1,1]$ range.  
The same preprocessing is applied to both training and test images.

\paragraph{Normalization.}
Across all datasets, images are scaled to $[-1,1]$, which matches the input
range expected by the U\hspace{0.15mm}-Net denoisers used in DDPM/DM/SDXL models.
We avoid augmentation because diffusion inversion and semantic editing require
consistent reconstruction of the original image.

\subsection{Metrics}

\paragraph{Editing Precision Ratio (EPR).}
For EPR, we use the classifier \emph{logits} rather than probabilities, as logits
preserve relative semantic strength without saturation effects.  
We set $\epsilon = 10^{-8}$ purely for numerical stability; this does not affect
the scale or ranking of the metric.  
For each target attribute $c$, all remaining annotated attributes in the dataset
are treated as non-target attributes.  
This design evaluates unintended changes across the full attribute space, and
datasets with rich attribute annotations (e.g., CelebA-HQ with 40+ labels)
provide more reliable and discriminative EPR estimates.

\paragraph{Learned Perceptual Image Patch Similarity (LPIPS).}
We evaluate perceptual distortion using LPIPS~\cite{zhang2018unreasonable}.  
Following the official implementation, we adopt the AlexNet-based LPIPS model
(\texttt{lpips.LPIPS(net='alex')} in our code).  
All images are resized to the target resolution and normalized to the 
$[-1,1]$ range before evaluation.  
Given an original image $x$ and its edited counterpart $\hat{x}$, LPIPS computes 
the $L_2$ distance between deep features extracted from a pretrained 
perceptual network.  
Lower LPIPS indicates better preservation of non-target visual details.

\paragraph{CLIP-Score.}
To measure the semantic alignment between the edited images and the target
attribute, we compute CLIP-Score using the ViT-L/14 CLIP model.  
For a text prompt $t$ describing the target attribute, the CLIP encoder extracts
a normalized text representation $f_t$, and each edited image $\hat{x}$ is mapped
to a normalized visual feature $f_{\hat{x}}$.  
The CLIP-Score is the cosine similarity
\[
\mathrm{CLIP}(\hat{x}, t)
= \left\langle f_{\hat{x}},\, f_t \right\rangle,
\]
where higher scores indicate stronger semantic consistency with the target
attribute.  
All images are normalized using the default CLIP preprocessing pipeline, and
scores are averaged over the entire evaluation set.

\paragraph{ArcFace Similarity.}
To quantify identity preservation in face-editing experiments, we compute the 
cosine similarity between ArcFace embeddings of the original and edited images.  
Following the standard InsightFace implementation~\cite{deng2019arcface}, each 
face is detected and aligned using the SCRFD detector, and a 512-dimensional 
normalized embedding is extracted by the \texttt{buffalo\_l} model (ResNet-50 
backbone).  

Given two embeddings $e, \hat{e} \in \mathbb{R}^{512}$, identity similarity is
\[
\mathrm{Sim}(e,\hat{e})
= \frac{\langle e,\hat{e}\rangle}{\|e\|_2 \,\|\hat{e}\|_2}
\in [0,1],
\]
where higher values indicate stronger identity consistency.  
Images are evaluated after automatic face alignment, and pairs with undetected
faces are excluded, following common practice in prior work.

\subsection{Baseline Configuration}
\subsubsection{Concept Sliders}
We implement Concept Sliders~\cite{2024concept} as our baseline method for controllable image editing. The baseline configuration follows the original implementation with the following specifications:

\paragraph{Model Architecture.}
We use Stable Diffusion v1.4 as the base diffusion model with LoRA adaptation for concept manipulation. The LoRA network employs a \texttt{c3lier} architecture targeting both linear and convolutional layers in the UNet, with rank $r=4$ and scaling factor $\alpha=1.0$. Training is performed using the \texttt{noxattn} method, excluding cross-attention layers from adaptation.

\paragraph{Training Configuration.}
The model is trained for 1,000 iterations using the AdamW optimizer with a constant learning rate of $2 \times 10^{-4}$. We use DDIM scheduling with 50 denoising steps and classifier-free guidance scale of 4.0. All training is conducted at $512 \times 512$ resolution with bfloat16 precision.

\paragraph{Concept Definitions.}
For each target attribute, we define concept pairs using natural language prompts:
\begin{itemize}
    \item \textbf{Smiling}: ``smiling person'' vs. ``sad person''
    \item \textbf{Big Nose}: ``person with big nose'' vs. ``person with small nose''
    \item \textbf{Beards}: ``person with beards'' vs. ``person without beards''
    \item \textbf{Blonde Hair}: ``person with blonde hair'' vs. ``person without blonde hair''
    \item \textbf{Young}: ``very young person'' vs. ``very old person''
\end{itemize}
All concepts use ``person'' as the neutral anchor point for conditioning.

\paragraph{Inference Protocol.}
During inference, we apply the trained LoRA weights with editing strength $\alpha \in \{1, 32, 64, 96, 128\}$ using DDIM sampling with 50 steps. The LoRA scaling is activated after timestep 500 to preserve image fidelity while enabling concept manipulation. We use classifier-free guidance with scale 7.5 and the prompt ``a photo of a person'' for all evaluations.

\paragraph{Implementation Details.}
Models are implemented using the Diffusers library and trained on NVIDIA A100 GPUs. LoRA weights are saved every 500 iterations, with the final checkpoint used for evaluation. All images are processed at $512 \times 512$ resolution with standard normalization.

\subsubsection{Asyrp}
\label{sec:asyrp_baseline}

We implement Asyrp~\cite{kwon2022diffusion} as another baseline method for semantic image editing. Asyrp performs editing by learning delta blocks that modify intermediate features in the diffusion denoising process at specific timesteps.

\paragraph{Model Architecture.}
The method uses a pre-trained diffusion model on CelebA-HQ with a simple UNet architecture. The UNet consists of 128 base channels with channel multipliers $[1, 1, 2, 2, 4, 4]$, 2 residual blocks per resolution, and self-attention at 16×16 resolution. All processing is performed at $256 \times 256$ resolution with 3 input/output channels.

\paragraph{Diffusion Configuration.}
The diffusion process employs a linear noise schedule with $\beta_{\text{start}} = 10^{-4}$, $\beta_{\text{end}} = 0.02$, and $T = 1000$ timesteps. Sampling uses DDIM with 50 denoising steps during both training and inference.

\paragraph{Training Protocol.}
Delta blocks are trained using 100 training images and 32 test images over 5 iterations with batch size 1. The learning rate is set to 0.5 with a combined loss function:
\begin{equation}
\mathcal{L} = \lambda_{\text{CLIP}} \mathcal{L}_{\text{CLIP}} + \lambda_{L1} \mathcal{L}_{L1}
\end{equation}
where $\lambda_{\text{CLIP}} = 1.0$ and $\lambda_{L1} = 3.0$. Training employs 50 DDIM inversion steps ($n_{\text{inv}}$) and 50 generation steps ($n_{\text{train}}$).

\paragraph{Temporal Control.}
The method uses adaptive timestep selection based on LPIPS distance thresholds:
\begin{itemize}
    \item \textbf{Edit timestep} $t_{\text{edit}} = 500$: when delta blocks begin activation
    \item \textbf{LPIPS thresholds}: $\tau_{\text{edit}} = 0.33$, $\tau_{\text{noise}} = 0.1$ for automatic timestep determination
\end{itemize}

\paragraph{Concept Definitions.}
Semantic concepts are defined using text prompt pairs from source to target:
\begin{itemize}
    \item \textbf{Smiling}: ``face'' $\rightarrow$ ``smiling face''
    \item \textbf{Big Nose}: ``person'' $\rightarrow$ ``person with big nose''
    \item \textbf{Beards}: ``person'' $\rightarrow$ ``person with beards''
    \item \textbf{Blonde Hair}: ``person'' $\rightarrow$ ``person with blond hair''
    \item \textbf{Age}: ``person'' $\rightarrow$ ``young person'' / ``old person''
\end{itemize}

\paragraph{Inference Procedure.}
During inference, the method follows a three-stage process: (1) DDIM inversion with 50 steps to obtain latent codes, (2) forward diffusion with delta block injection starting at $t_{\text{edit}}$, and (3) optional noise injection from $t_{\text{noise}}$ using the \texttt{add\_noise\_from\_xt} strategy. The editing strength is controlled by the coefficient $\alpha = 1.0$.

\subsubsection{BoundaryDiffusion.}
We include BoundaryDiffusion~\cite{zhu2023boundary} as a representative
classifier-guided editing baseline.  
For each semantic attribute, we use the official CLIP-based boundaries provided
by the authors, which contain both activation-space boundaries 
($W_h$) and latent-space boundaries ($W_z$).
During editing, the U\hspace{0.1mm}-Net activation at timestep $t=500$ is perturbed
along the attribute direction by traversing the boundary over a fixed range
($[-500, 500]$ in our implementation), following the protocol in the original
work.  
All methods use the same DDIM inversion setup ($n_{\text{inv}}=40$ steps) and the
same number of forward editing steps ($n_{\text{edit}}=40$).  
The boundary is applied only once per denoising step, and all other hyperparameters
are kept identical to the official implementation for fair comparison.

\subsubsection{MasaCtrl}
We employ MasaCtrl~\cite{cao2023masactrl} as a baseline method. This approach achieves structure-preserving image editing by controlling the mutual self-attention mechanism within the U-Net architecture of diffusion models. Our configuration uses Stable Diffusion v1.4 as the pre-trained foundation model with a DDIM scheduler for the denoising process, setting beta\_start=0.00085, beta\_end=0.012, and beta\_schedule="scaled\_linear". The MasaCtrl control parameters are configured as follows: start\_step=4, start\_layer=10, total inference steps=50, and guidance\_scale=7.5. The method first performs DDIM inversion to map the source image into the latent space, then applies mutual self-attention control during specified denoising steps (4-50) and U-Net layers (10-16) to achieve semantic editing while preserving structural coherence. Image preprocessing includes resizing input images to 512×512 resolution and normalizing pixel values to the [-1,1] range. To ensure reproducibility, we set fixed random seeds for each image during the editing process.

\subsubsection{SwiftEdit:} 
We employ SwiftEdit as a baseline method for lightning-fast text-guided image editing via one-step diffusion. SwiftEdit utilizes a pre-trained diffusion model with single-step inversion and sampling to achieve significant efficiency improvements. Our configuration includes: image resolution resized to $512 \times 512$ pixels, middle timestep $t_{mid}=500$, and final timestep $t_{final}=999$. The mask generation parameters are set as follows: text hidden state scaling factor $\alpha_{ta}=1$, editing region scaling factor $\alpha_{edit}=0.2$, non-editing region scaling factor $\alpha_{non-edit}=1$, mask threshold $\tau_{mask}=0.5$, and clamp rate $\gamma_{clamp}=3.0$. The method first predicts inverted noise through the inverse U-Net, then estimates editing masks based on the difference between source and edit prompts, and finally applies mask controllers at specified U-Net layers ("mid\_blocks" and "up\_blocks") for localized editing.

\begin{table*}[h]
\centering
\small
\begin{tabular}{llccc}
\toprule
\textbf{Dataset} & \textbf{Concept} & $y^{\text{origin}}$ & $y^{\text{ref}}$ & $\lambda_{\text{CLIP}} : \lambda_{\text{recons}}$ \\
\midrule
\multirow{9}{*}{CelebA-HQ} 
  & Smiling            & face    & smiling face                    & 3:1 \\
  & Big Nose           & person  & person with big nose            & 3:1 \\
  & Blond Hair         & person  & person with blond hair          & 3:1 \\
  & Beards             & person  & person with beards              & 3:1 \\
  & Eyeglasses         & person  & person wearing eyeglasses       & 3:1 \\
  & Narrow eyes        & face    & face with narrow eyes           & 3:1 \\
  & arched eyebrow     & face    & face with arched eyebrows       & 3:1 \\
  & Young              & person  & young person                    & 3:1 \\
\midrule
\multirow{5}{*}{FFHQ} 
  & Beards             & person  & person with beards              & 3:1 \\
  & Young              & person  & young person                    & 3:1 \\
  & tanned             & face    & tanned face                     & 3:1 \\
  & angry              & face    & angry face                      & 3:1 \\
  & sad                & face    & sad face                        & 3:1 \\
\midrule
\multirow{4}{*}{LSUN-Church} 
  & Gothic             & Church  & Gothic Church                   & 2.5:1 \\
  & Red Brick          & Church  & Red brick wall Church           & 2.5:1 \\
  & Snowy              & Church  & snowy day                       & 2.5:1 \\
  & wooden             & Church  & Wooden House                    & 2.5:1 \\
\midrule
\multirow{4}{*}{AFHQ-Dog} 
  & Angry              & Dog     & Angry Dog                       & 2.2:1 \\
  & Sleepy             & Dog     & Sleepy Dog                      & 2.2:1 \\
  & Smiling            & Dog     & Smiling Dog                     & 2.2:1 \\
  & puppy              & Dog     & Puppy                            & 2.2:1 \\
\bottomrule
\end{tabular}
\caption{Hyperparameter settings for CLIP loss and reconstruction loss for different datasets and concepts.}
\label{tab:clip_hyper}
\end{table*}

\begin{table}[ht]\footnotesize
\centering
\setlength{\tabcolsep}{4pt}
\begin{tabular}{lccc}
\toprule
\textbf{Params} & \textbf{MSE} $\downarrow$ & \textbf{Cosine Sim.} $\uparrow$ & \textbf{DAR ($\tau=0.01$)} $\downarrow$ \\
\midrule
$\lambda_{\text{sparse}}=0.5$& $2.34 \times 10^{-2}$ & 0.993 & $2.72 \times 10^{-2}$  \\
$\lambda_{\text{sparse}}=1$& $2.33 \times 10^{-2}$ & 0.994 & $2.50 \times 10^{-2}$  \\
$\lambda_{\text{sparse}}=2$& $2.18 \times 10^{-2}$ & 0.994 & $2.23 \times 10^{-2}$  \\
$\lambda_{\text{sparse}}=4$& $2.07 \times 10^{-2}$ & 0.994 & $2.03 \times 10^{-2}$  \\
$\lambda_{\text{sparse}}=8$& $2.30 \times 10^{-2}$ & 0.993 & $1.80 \times 10^{-2}$  \\
$\lambda_{\text{sparse}}=16$& $2.77 \times 10^{-1}$ & 0.987 & $4.81 \times 10^{-2}$  \\
$\lambda_{\text{sparse}}=32$& $3.69 \times 10^{-1}$ & 0.986 & $1.72 \times 10^{-2}$  \\

\bottomrule
\end{tabular}
\caption{Reconstruction quality and sparsity of the SAE trained on FFHQ across different sparsity loss weight $\lambda_{\text{sparse}}$.}

\label{tab:sae_eval}
\end{table}
\subsection{Choice of $\alpha$ in Different Concepts}
The hyperparameter $\alpha$ determines the editing intensity along each semantic direction. In our experiments, the value of $\alpha$ for each concept is primarily chosen based on visual inspection of the editing results on the validation set. Specifically, we adjust $\alpha$ to achieve a balance between perceptible attribute change and preservation of image realism. Notably, for attributes that are abundant in the dataset (e.g., \textit{smiling} in CelebA-HQ), a smaller $\alpha$ is typically sufficient to induce clear semantic changes, whereas less common or more subtle attributes often require a larger value of $\alpha$. The final $\alpha$ settings for all datasets and concepts are summarized in Table~\ref{tab:alpha_selection}.

We do not rely on the Editing Precision Ratio (EPR) metric for the selection of $\alpha$ due to two primary limitations. First, EPR can only be computed on datasets with explicit attribute annotations (e.g., CelebA-HQ), restricting its applicability to many datasets or concepts without such labels. Second, EPR is inherently limited to the predefined set of annotated attributes, and therefore fails to capture editing quality for concepts or semantic directions that are not covered by existing labels. In contrast, visual inspection enables a more comprehensive and flexible assessment of editing performance across diverse and potentially unannotated attributes.

\begin{table}[h]
\centering
\small
\resizebox{!}{!}{
\begin{tabular}{llccc}
\toprule
\textbf{Dataset}& Concept& $\alpha$ \\
\midrule
\multirow{9}{*}{CelebA-HQ} 
  & Smiling            & 128 \\
  & Big Nose           & 64 \\
  & Blond Hair         & 128 \\
  & Beards             & 96 \\
  & Eyeglasses         & 96 \\
  & Narrow eyes        & 160 \\
  & arched eyebrow     & 160 \\
  & Young              & 128 \\
\midrule
\multirow{5}{*}{FFHQ} 
  & Beards             & 96 \\
  & Young              & 128 \\
  & tanned             & 80 \\
  & angry              & 80 \\
  & sad                & 64 \\
\midrule
\multirow{4}{*}{LSUN-Church} 
  & Gothic             & 16 \\
  & Red Brick          & 16 \\
  & Snowy              & 16 \\
  & wooden             & 16 \\
\midrule
\multirow{4}{*}{AFHQ-Dog} 
  & Angry              & 32 \\
  & Sleepy             & 64 \\
  & Smiling            & 31 \\
  & puppy              & 64 \\
\bottomrule
\end{tabular}
}
\caption{Hyperparameter settings for $\alpha$ for different datasets and concepts.}
\label{tab:alpha_selection}
\end{table}

\section{User Study Design}
\label{sec:user_study}

To further evaluate the effectiveness of our Editing Precision Ratio (EPR) metric, we conducted a user study to assess whether the metric aligns with human perception of editing quality.

\subsection{Experimental Setup}

In this experiment, we invited \textbf{10} participants to rate a collection of edited images generated by different semantic editing methods, including ours and baseline approaches. Each test sample consists of the following:

\begin{itemize}
    \item \textbf{Original Image:} The unedited reference image.
    \item \textbf{Target Description:} The semantic attribute to be edited (e.g., ``smiling'').
    \item \textbf{Edited Image:} The result of semantic editing according to the target attribute.
\end{itemize}

For each sample, participants were instructed to evaluate the edited image on two aspects:

\begin{enumerate}
    \item \textbf{Target Change:} How much does the edited image match the target description?\\
    (1 = No visible change, 10 = Completely matches the target description)
    \item \textbf{Identity Preservation:} How well does the edited image preserve the identity of the original image?\\
    (1 = Perfectly preserves identity, 10 = Looks like a completely different person/image)
\end{enumerate}

Participants were asked to provide their ratings in the format: \textbf{Target/Non-target} (e.g., 3/2).

\subsection{Procedure}

The method names and editing settings were anonymized to ensure a fair assessment. Prior to the formal study, a brief tutorial and several examples were given to each participant.

\subsection{Participants}

A total of \textbf{10} participants were recruited for the user study. (Details of participant demographics are omitted for anonymity.)

\subsection{Results}

Detailed results are provided in the supplementary file \texttt{Human\_Evaluation.pdf}. We further observe that the trends in human evaluation are largely aligned with those measured by the EPR metric, demonstrating the rationality and practical effectiveness of our approach.

\subsection{Summary}

This user study aims to assess the subjective perception of editing precision and identity preservation, and to validate the consistency between the EPR metric and human judgment.

\section{Performance of the Attribute Classifier}
To ensure fair and robust evaluation, we independently trained four attribute classifiers, ResNet18, VGG16, MobileNetV2, and ViT-B/16, using identical settings on the \textbf{cropped CelebA} dataset. All classifiers are fine-tuned from ImageNet-pretrained weights.

Following the standard protocol, we split the dataset as follows: images with indices \texttt{image\_n < 162771} are used for training, \texttt{162771 $\leq$ image\_n < 182638} for validation, and \texttt{image\_n $\geq$ 182638} for testing.

\begin{itemize}
    \item \textbf{ResNet18} is a classical convolutional neural network (CNN) featuring residual connections, which enable effective training of deep models by mitigating the vanishing gradient problem.
    \item \textbf{VGG16} is a deep CNN characterized by its uniform use of $3\times3$ convolutional layers and simple, sequential architecture, widely adopted as a strong baseline for visual tasks.
    \item \textbf{MobileNetV2} is a lightweight CNN designed for efficient inference on mobile and embedded devices, employing depthwise separable convolutions and inverted residual blocks to reduce computational cost while maintaining competitive performance.
    \item \textbf{ViT-B/16} is a vision transformer model that processes images as sequences of non-overlapping patches and leverages self-attention mechanisms to capture global dependencies, marking a departure from traditional CNN-based designs.
\end{itemize}

The tables below(table.~\ref{tab:celeba_classifier}, table.~\ref{tab:vit_classifier_perf}, table.~\ref{tab:celeba_mobilenetv2}, table.~\ref{tab:vgg_classifier_perf}) reports their performance on the test set. As shown, all classifiers achieve consistently high accuracy, demonstrating stable and reliable attribute prediction across different initializations.

\subsection{Ablation Study on EPR Sensitivity}

To better understand the behavior and limitations of the Editing Precision Ratio (EPR), 
we conduct an ablation study using different attribute classifiers, including 
ResNet18, MobileNetV2, VGG16, and ViT-B/16.  
The results are summarized in Table~\ref{tab:epr_full}, which reports EPR for 
five facial attributes across all editing methods.

Overall, EPR values exhibit substantial variation across classifier backbones.  
For example, the same edited images can yield noticeably different EPR scores when 
evaluated by ResNet18 versus VGG16, and ViT-B/16 shows an even larger degree of 
inconsistency, often producing unstable or uninformative scores.  
This divergence indicates that EPR is highly sensitive to the choice of attribute  
classifier: models with different architectures, feature embeddings, and decision 
boundaries respond differently to subtle attribute shifts introduced during editing.

While EPR provides a convenient quantitative measure of editing specificity under 
supervised attribute labels, this ablation highlights several inherent limitations.  
First, EPR fundamentally depends on high-quality, concept-aligned labels, and cannot be 
reliably computed when labels are noisy, ambiguous, or unavailable.  
Second, its dependence on a particular classifier introduces model-induced bias, 
making absolute EPR values difficult to compare across backbones or datasets.  
Finally, concepts with broader visual variation or weak label definitions tend to 
produce more volatile EPR scores, limiting the metric's robustness in open-domain 
or attribute-sparse settings.

Taken together, these findings suggest that EPR should be interpreted with caution and 
primarily as a relative metric within a fixed classifier setting, rather than as an 
absolute measure of editing quality.  
This motivates the need for complementary evaluation metrics, such as  
CLIP-Score and LPIPS, that do not rely on explicit attribute labels or classifier 
decision boundaries.

\begin{table*}[ht]
\centering
\begin{tabular}{lccccc}
\toprule
Attribute & Accuracy & Balanced Acc & F1 & Precision & Recall \\
\midrule
5\_o\_Clock\_Shadow & 94.823 & 86.954 & 74.849 & 72.718 & 77.121 \\
Arched\_Eyebrows & 84.358 & 80.833 & 72.543 & 72.449 & 72.656 \\
Attractive & 83.173 & 83.163 & 82.846 & 83.766 & 81.962 \\
Bags\_Under\_Eyes & 85.504 & 78.772 & 65.342 & 63.375 & 67.451 \\
Bald & 99.067 & 86.849 & 77.095 & 80.376 & 74.090 \\
Bangs & 96.204 & 91.944 & 87.557 & 89.435 & 85.757 \\
Big\_Lips & 72.832 & 62.683 & 44.509 & 67.068 & 33.346 \\
Big\_Nose & 84.297 & 76.823 & 63.289 & 62.761 & 63.847 \\
Black\_Hair & 90.484 & 86.472 & 81.599 & 85.930 & 77.687 \\
Blond\_Hair & 96.182 & 91.368 & 85.547 & 86.308 & 84.805 \\
Blurry & 96.380 & 73.442 & 57.232 & 71.188 & 47.921 \\
Brown\_Hair & 89.496 & 83.691 & 71.858 & 69.291 & 74.631 \\
Bushy\_Eyebrows & 93.038 & 80.330 & 70.158 & 78.885 & 63.179 \\
Chubby & 95.942 & 75.738 & 58.131 & 64.195 & 53.138 \\
Double\_Chin & 96.470 & 72.727 & 54.690 & 66.250 & 46.594 \\
Eyeglasses & 99.684 & 98.401 & 97.541 & 98.162 & 96.928 \\
Goatee & 97.593 & 88.127 & 74.747 & 72.023 & 77.705 \\
Gray\_Hair & 98.296 & 84.933 & 72.536 & 74.555 & 70.660 \\
Heavy\_Makeup & 92.040 & 91.481 & 90.008 & 91.535 & 88.538 \\
High\_Cheekbones & 88.085 & 87.996 & 87.370 & 89.293 & 85.529 \\
Male & 98.373 & 98.214 & 97.887 & 98.261 & 97.517 \\
Mouth\_Slightly\_Open & 94.339 & 94.332 & 94.241 & 94.938 & 93.555 \\
Mustache & 97.112 & 72.738 & 55.364 & 68.804 & 46.321 \\
Narrow\_Eyes & 87.750 & 64.538 & 43.325 & 69.442 & 31.503 \\
No\_Beard & 96.516 & 93.404 & 97.956 & 98.109 & 97.804 \\
Oval\_Face & 76.107 & 63.041 & 43.454 & 72.329 & 31.079 \\
Pale\_Skin & 97.200 & 75.297 & 60.674 & 74.176 & 51.381 \\
Pointy\_Nose & 77.812 & 68.238 & 54.168 & 66.095 & 45.898 \\
Receding\_Hairline & 94.025 & 74.910 & 59.570 & 69.943 & 51.889 \\
Rosy\_Cheeks & 95.408 & 77.371 & 63.751 & 73.512 & 56.313 \\
Sideburns & 97.952 & 89.587 & 78.453 & 76.631 & 80.367 \\
Smiling & 93.365 & 93.366 & 93.268 & 94.721 & 91.859 \\
Straight\_Hair & 84.999 & 74.687 & 61.427 & 66.734 & 56.912 \\
Wavy\_Hair & 85.321 & 82.027 & 77.616 & 87.235 & 69.911 \\
Wearing\_Earrings & 90.660 & 86.182 & 77.657 & 76.787 & 78.550 \\
Wearing\_Hat & 99.176 & 94.328 & 90.086 & 91.165 & 89.035 \\
Wearing\_Lipstick & 94.172 & 94.236 & 94.322 & 95.931 & 92.768 \\
Wearing\_Necklace & 88.140 & 65.181 & 43.759 & 63.258 & 33.476 \\
Wearing\_Necktie & 97.088 & 87.067 & 78.399 & 81.664 & 75.411 \\
Young & 89.021 & 82.631 & 92.913 & 90.866 & 95.056 \\
\midrule
Cumulative avg & 91.712 & 86.739 & 81.218 & 85.327 & 77.487 \\
Attr avg mean & 91.712 & 82.353 & 73.193 & 78.754 & 69.754 \\
\bottomrule
\end{tabular}
\caption{Performance of ResNet-18 attribute classifier on CelebA (test set).}
\label{tab:celeba_classifier}
\end{table*}

\begin{table*}[ht]
\centering
\begin{tabular}{lrrrrr}
\toprule
          Attribute &  Accuracy &  Balanced Acc &     F1 &  Precision &  Recall \\
\midrule
   5\_o\_Clock\_Shadow &    91.705 &        78.454 & 63.491 &     66.067 &  61.109 \\
    Arched\_Eyebrows &    80.717 &        75.203 & 63.096 &     62.417 &  63.790 \\
         Attractive &    77.224 &        77.149 & 78.298 &     77.602 &  79.007 \\
    Bags\_Under\_Eyes &    80.717 &        68.110 & 50.046 &     54.087 &  46.566 \\
               Bald &    98.374 &        73.924 & 55.201 &     64.194 &  48.418 \\
              Bangs &    94.166 &        87.406 & 79.656 &     81.560 &  77.839 \\
           Big\_Lips &    78.140 &        60.898 & 33.563 &     31.406 &  36.038 \\
           Big\_Nose &    79.061 &        68.338 & 52.759 &     60.135 &  46.996 \\
         Black\_Hair &    89.681 &        85.422 & 75.950 &     73.904 &  78.113 \\
         Blond\_Hair &    94.458 &        88.492 & 81.598 &     83.396 &  79.876 \\
             Blurry &    94.956 &        64.091 & 36.015 &     45.048 &  30.000 \\
         Brown\_Hair &    83.843 &        75.279 & 63.688 &     69.558 &  58.731 \\
        Bushy\_Eyebrows &    90.985 &        78.912 & 66.227 &     71.036 &  62.028 \\
             Chubby &    94.196 &        66.582 & 42.551 &     53.982 &  35.115 \\
        Double\_Chin &    95.520 &        68.609 & 45.930 &     56.334 &  38.769 \\
          Eyeglasses &    98.102 &        90.551 & 85.714 &     90.048 &  81.779 \\
             Goatee &    95.012 &        78.792 & 63.845 &     68.520 &  59.768 \\
           Gray\_Hair &    97.448 &        82.861 & 71.786 &     77.711 &  66.701 \\
       Heavy\_Makeup &    89.435 &        88.979 & 86.520 &     86.138 &  86.905 \\
    High\_Cheekbones &    85.755 &        85.560 & 84.065 &     84.503 &  83.632 \\
               Male &    97.443 &        97.404 & 97.001 &     96.864 &  97.139 \\
Mouth\_Slightly\_Open &    91.307 &        91.292 & 90.969 &     91.079 &  90.860 \\
           Mustache &    95.435 &        68.332 & 45.786 &     57.164 &  38.185 \\
        Narrow\_Eyes &    91.101 &        68.001 & 40.790 &     40.763 &  40.818 \\
           No\_Beard &    93.104 &        87.510 & 95.823 &     95.462 &  96.187 \\
          Oval\_Face &    71.702 &        61.514 & 43.155 &     49.341 &  38.347 \\
          Pale\_Skin &    96.341 &        71.982 & 51.630 &     59.969 &  45.327 \\
        Pointy\_Nose &    72.769 &        63.140 & 46.030 &     52.864 &  40.760 \\
 Receding\_Hairline &    93.321 &        71.419 & 49.678 &     54.222 &  45.836 \\
        Rosy\_Cheeks &    94.438 &        73.303 & 54.545 &     61.789 &  48.822 \\
          Sideburns &    95.616 &        78.372 & 64.694 &     72.545 &  58.376 \\
            Smiling &    90.915 &        90.872 & 90.504 &     91.443 &  89.586 \\
      Straight\_Hair &    79.856 &        66.129 & 46.640 &     51.215 &  42.815 \\
          Wavy\_Hair &    80.284 &        77.555 & 66.718 &     62.576 &  71.447 \\
   Wearing\_Earrings &    84.945 &        73.431 & 58.139 &     61.889 &  54.817 \\
         Wearing\_Hat &    98.294 &        89.652 & 81.626 &     83.204 &  80.106 \\
     Wearing\_Lipstick &    91.136 &        91.412 & 90.435 &     87.164 &  93.962 \\
   Wearing\_Necklace &    86.420 &        58.067 & 26.883 &     38.331 &  20.701 \\
    Wearing\_Necktie &    95.319 &        79.017 & 65.038 &     71.076 &  59.945 \\
               Young &    84.728 &        76.719 & 90.088 &     87.388 &  92.961 \\
\midrule
    Cumulative avg &    89.349 &        83.781 & 75.443 &     77.189 &  73.774 \\
    Attr avg mean &    89.349 &        76.968 & 64.404 &     68.100 &  61.704 \\
\bottomrule
\end{tabular}
\caption{Performance of ViT-B/16 attribute classifier on CelebA (test set).}
\label{tab:vit_classifier_perf}
\end{table*}

\begin{table*}[ht]
\centering
\begin{tabular}{lccccc}
\toprule
Attribute & Accuracy & Balanced Acc & F1 & Precision & Recall \\
\midrule
5\_o\_Clock\_Shadow & 94.357 & 84.224 & 74.803 & 79.087 & 70.959 \\
Arched\_Eyebrows & 86.339 & 82.280 & 73.650 & 73.422 & 73.880 \\
Attractive & 81.985 & 81.930 & 82.786 & 82.283 & 83.295 \\
Bags\_Under\_Eyes & 84.240 & 73.585 & 59.311 & 63.850 & 55.375 \\
Bald & 99.018 & 83.899 & 74.172 & 81.395 & 68.127 \\
Bangs & 95.898 & 91.404 & 85.883 & 86.739 & 85.043 \\
Big\_Lips & 82.020 & 65.167 & 41.056 & 41.247 & 40.867 \\
Big\_Nose & 83.480 & 74.608 & 63.173 & 70.925 & 56.949 \\
Black\_Hair & 91.710 & 87.876 & 80.358 & 79.439 & 81.298 \\
Blond\_Hair & 95.777 & 91.226 & 86.047 & 87.487 & 84.653 \\
Blurry & 96.653 & 74.839 & 58.925 & 70.250 & 50.745 \\
Brown\_Hair & 86.067 & 78.005 & 68.373 & 75.575 & 62.424 \\
Bushy\_Eyebrows & 92.742 & 81.615 & 72.173 & 79.541 & 66.054 \\
Chubby & 95.767 & 74.029 & 58.754 & 72.783 & 49.260 \\
Double\_Chin & 96.703 & 74.484 & 59.742 & 74.540 & 49.846 \\
Eyeglasses & 99.577 & 97.967 & 96.937 & 97.792 & 96.095 \\
Goatee & 96.869 & 86.584 & 77.817 & 81.418 & 74.522 \\
Gray\_Hair & 97.966 & 85.636 & 77.506 & 83.957 & 71.975 \\
Heavy\_Makeup & 92.576 & 92.321 & 90.549 & 89.944 & 91.162 \\
High\_Cheekbones & 88.312 & 88.107 & 86.874 & 87.677 & 86.086 \\
Male & 98.812 & 98.791 & 98.606 & 98.559 & 98.652 \\
Mouth\_Slightly\_Open & 94.151 & 94.138 & 93.922 & 94.060 & 93.785 \\
Mustache & 96.275 & 73.778 & 56.927 & 68.392 & 48.754 \\
Narrow\_Eyes & 93.467 & 74.822 & 54.868 & 57.009 & 52.882 \\
No\_Beard & 96.215 & 93.322 & 97.701 & 97.594 & 97.809 \\
Oval\_Face & 75.940 & 64.326 & 46.889 & 61.426 & 37.916 \\
Pale\_Skin & 96.909 & 72.726 & 56.330 & 72.000 & 46.262 \\
Pointy\_Nose & 77.324 & 68.244 & 54.222 & 63.812 & 47.138 \\
Receding\_Hairline & 94.604 & 76.338 & 59.455 & 64.691 & 55.003 \\
Rosy\_Cheeks & 95.248 & 74.830 & 59.554 & 71.209 & 51.178 \\
Sideburns & 97.347 & 87.059 & 79.582 & 84.596 & 75.128 \\
Smiling & 93.270 & 93.230 & 92.967 & 93.921 & 92.033 \\
Straight\_Hair & 85.423 & 73.915 & 60.534 & 68.275 & 54.370 \\
Wavy\_Hair & 86.661 & 85.155 & 77.230 & 73.156 & 81.783 \\
Wearing\_Earrings & 91.800 & 84.555 & 77.214 & 82.143 & 72.842 \\
Wearing\_Hat & 99.034 & 94.286 & 89.711 & 90.389 & 89.043 \\
Wearing\_Lipstick & 92.948 & 93.290 & 92.424 & 88.716 & 96.456 \\
Wearing\_Necklace & 89.545 & 65.786 & 44.301 & 61.965 & 34.474 \\
Wearing\_Necktie & 96.879 & 85.543 & 77.088 & 82.581 & 72.280 \\
Young & 88.237 & 81.305 & 92.369 & 89.559 & 95.361 \\
\midrule
Cumulative avg & 91.954 & 87.354 & 81.342 & 83.727 & 79.088 \\
Attr avg mean & 91.954 & 82.131 & 73.270 & 78.085 & 69.794 \\
\bottomrule
\end{tabular}
\caption{Performance of MobileNetV2 attribute classifier on CelebA (test set).}
\label{tab:celeba_mobilenetv2}
\end{table*}
\begin{table*}[ht]
\centering
\begin{tabular}{lrrrrr}
\toprule
          Attribute &  Accuracy &  Balanced Acc &     F1 &  Precision &  Recall \\
\midrule
   5\_o\_Clock\_Shadow &    94.378 &        86.009 & 75.911 &     76.789 &  75.053 \\
    Arched\_Eyebrows &    86.636 &        83.007 & 74.488 &     73.507 &  75.497 \\
         Attractive &    81.980 &        81.878 & 82.973 &     81.569 &  84.427 \\
    Bags\_Under\_Eyes &    84.633 &        74.648 & 60.854 &     64.519 &  57.583 \\
               Bald &    99.039 &        88.911 & 77.126 &     75.943 &  78.346 \\
              Bangs &    96.099 &        92.587 & 86.826 &     86.051 &  87.616 \\
           Big\_Lips &    84.799 &        64.574 & 41.654 &     50.563 &  35.414 \\
           Big\_Nose &    83.500 &        74.060 & 62.503 &     71.914 &  55.270 \\
         Black\_Hair &    91.856 &        89.017 & 81.169 &     78.395 &  84.146 \\
         Blond\_Hair &    95.666 &        92.138 & 86.070 &     85.120 &  87.042 \\
             Blurry &    96.688 &        77.385 & 61.565 &     68.264 &  56.064 \\
         Brown\_Hair &    86.228 &        79.327 & 69.808 &     74.092 &  65.992 \\
        Bushy\_Eyebrows &    92.898 &        83.223 & 73.661 &     78.108 &  69.693 \\
             Chubby &    95.938 &        75.927 & 61.553 &     73.160 &  53.125 \\
        Double\_Chin &    96.814 &        77.071 & 62.961 &     73.297 &  55.179 \\
          Eyeglasses &    99.562 &        98.159 & 96.844 &     97.162 &  96.529 \\
             Goatee &    96.920 &        88.906 & 79.184 &     78.862 &  79.508 \\
           Gray\_Hair &    98.117 &        88.561 & 80.128 &     82.404 &  77.973 \\
       Heavy\_Makeup &    92.767 &        92.692 & 90.878 &     89.453 &  92.349 \\
    High\_Cheekbones &    88.544 &        88.360 & 87.160 &     87.784 &  86.545 \\
               Male &    98.822 &        98.806 & 98.618 &     98.536 &  98.699 \\
Mouth\_Slightly\_Open &    94.403 &        94.382 & 94.170 &     94.527 &  93.816 \\
           Mustache &    96.371 &        76.141 & 59.878 &     67.758 &  53.639 \\
        Narrow\_Eyes &    93.532 &        78.983 & 58.959 &     56.315 &  61.863 \\
           No\_Beard &    96.421 &        93.647 & 97.827 &     97.704 &  97.950 \\
          Oval\_Face &    76.408 &        64.036 & 46.021 &     64.080 &  35.903 \\
          Pale\_Skin &    96.950 &        75.369 & 59.383 &     69.654 &  51.752 \\
        Pointy\_Nose &    77.425 &        68.240 & 54.202 &     64.215 &  46.890 \\
 Receding\_Hairline &    94.770 &        77.525 & 61.217 &     65.600 &  57.383 \\
        Rosy\_Cheeks &    95.138 &        77.466 & 61.575 &     66.955 &  56.996 \\
          Sideburns &    97.297 &        89.369 & 80.322 &     80.470 &  80.176 \\
            Smiling &    93.628 &        93.586 & 93.336 &     94.359 &  92.335 \\
      Straight\_Hair &    85.458 &        75.652 & 62.524 &     66.501 &  58.996 \\
          Wavy\_Hair &    86.772 &        85.850 & 77.797 &     72.607 &  83.785 \\
   Wearing\_Earrings &    92.203 &        85.752 & 78.655 &     82.295 &  75.323 \\
         Wearing\_Hat &    99.099 &        95.382 & 90.554 &     89.843 &  91.277 \\
     Wearing\_Lipstick &    92.676 &        93.035 & 92.147 &     88.292 &  96.354 \\
   Wearing\_Necklace &    89.480 &        63.894 & 40.894 &     63.421 &  30.175 \\
    Wearing\_Necktie &    96.703 &        85.544 & 76.156 &     80.215 &  72.488 \\
               Young &    88.473 &        81.470 & 92.533 &     89.595 &  95.672 \\
\midrule
    Cumulative avg &    92.127 &        87.838 & 81.865 &     83.678 &  80.130 \\
    Attr avg mean &    92.127 &        83.264 & 74.252 &     77.497 &  72.121 \\
\bottomrule
\end{tabular}
\caption{Performance of VGG16 attribute classifier on CelebA (test set).}
\label{tab:vgg_classifier_perf}
\end{table*}
\section{Additional Results}
\subsection{Quantitative Results}


To further examine the generalization ability of our method beyond human facial attributes, 
we additionally evaluate it on two structurally different datasets: LSUN-Church and 
AFHQ-Dog. 
For LSUN-Church, we consider four architectural concepts (Gothic, Red Brick, Snowy, Wooden), 
while AFHQ-Dog includes four canine appearance and expression traits (Angry, Sleepy, 
Smiling, Puppy-like). 
The corresponding quantitative comparisons are presented in 
Table~\ref{tab:statistic_result_all} and Table~\ref{tab:statistic_result_dog}. 
Across both datasets, our approach produces stable and semantically aligned edits while 
maintaining distortion levels comparable to, or lower than, existing editing baselines. 
These observations suggest that the sparse latent directions learned by our method 
capture concept-level semantics that transfer reasonably well across diverse visual domains.

It is important to note that we do not report EPR or ArcFace on these datasets. 
EPR is specifically designed for settings where each concept is associated with 
well-defined attribute labels; however, LSUN-Church and AFHQ-Dog do not provide 
reliable per-image attribute annotations, making EPR ill-defined in these domains. 
Furthermore, ArcFace is a face recognition model and does not apply to non-human 
subjects such as buildings or animals. 
We therefore limit our evaluation to CLIP-Score and LPIPS, which remain applicable 
and provide a consistent measure of semantic alignment and perceptual distortion 
across a broader range of concepts.

While EPR offers a unified way to quantify the balance between editing strength and 
unintended attribute changes, its reliance on explicit attribute labels is a fundamental 
limitation. 
In label-scarce or open-domain scenarios, such as LSUN-Church and AFHQ-Dog, 
EPR cannot be meaningfully computed. 
These results highlight both the usefulness and the scope constraints of label-based 
metrics, and motivate future work toward developing label-free editing precision metrics 
that remain valid across heterogeneous visual domains.

\begin{table*}[ht]
\centering
\renewcommand{\arraystretch}{1.15}
\definecolor{mygray}{gray}{0.92}

\newcommand{\pmgray}[1]{\textcolor{lightgray}{\text{\scriptsize$\pm$#1}}}
\resizebox{\textwidth}{!}{
\begin{tabular}{llccccccccccccc}
\toprule
\textbf{Model} & \textbf{Concept} & \textbf{recon.} &
\textbf{Boundary}~\cite{zhu2023boundary} &
\textbf{Asyrp}~\cite{kwon2022diffusion} &
\textbf{Slider}~\cite{2024concept} &
\textbf{MasaCtrl}~\cite{cao2023masactrl} &
\textbf{SwiftEdit}~\cite{Nguyen_2025_CVPR} &
\textbf{CASL-Steer} \\
\midrule
\multirow{5}{*}{ResNet18} 
& Smiling
& 0.679\pmgray{0.289} 
& 1.231\pmgray{0.976} 
& 3.199\pmgray{2.182}
& 0.949\pmgray{0.479}
& 1.685\pmgray{1.375}
& 3.359\pmgray{2.064}
& \cellcolor{mygray}4.465\pmgray{2.319} \\

& Big Nose
& 0.207\pmgray{0.280}
& 0.882\pmgray{0.539}
& 0.938\pmgray{0.608}
& 1.035\pmgray{0.775}
& 0.793\pmgray{0.600}
& 0.867\pmgray{0.627}
& \cellcolor{mygray}1.060\pmgray{0.608} \\

& Young
& 0.773\pmgray{0.619}
& 1.411\pmgray{0.909}
& 1.685\pmgray{1.220}
& 0.937\pmgray{0.770}
& 1.427\pmgray{0.709}
& 1.536\pmgray{0.989}
& \cellcolor{mygray}1.817\pmgray{1.117} \\

& Beards
& 0.763\pmgray{0.669}
& 1.165\pmgray{0.974}
& 1.739\pmgray{1.268}
& 0.919\pmgray{0.383}
& 2.059\pmgray{1.541}
& 1.255\pmgray{0.884}
& \cellcolor{mygray}2.161\pmgray{1.080} \\

& Blond Hair
& 0.110\pmgray{0.032}
& 0.876\pmgray{0.794}
& 1.173\pmgray{1.002}
& 0.760\pmgray{0.548}
& 1.173\pmgray{0.678}
& 1.302\pmgray{1.014}
& \cellcolor{mygray}1.386\pmgray{1.213} \\

\midrule
\multirow{5}{*}{MobileNetV2}
& Smiling
& 0.848\pmgray{0.375}
& 1.292\pmgray{1.029}
& 3.490\pmgray{3.346}
& 1.326\pmgray{0.965}
& 1.668\pmgray{1.312}
& 4.516\pmgray{1.814}
& \cellcolor{mygray}3.237\pmgray{1.803}\\
& Big Nose
& 0.169\pmgray{0.081}
& 0.995\pmgray{0.423}
& 0.872\pmgray{0.292}
& 1.381\pmgray{0.820}
& 0.759\pmgray{0.555}
& 1.045\pmgray{0.749}
& \cellcolor{mygray}1.261\pmgray{0.650}\\
& Young
& 0.799\pmgray{0.571}
& 1.532\pmgray{1.099}
& 1.808\pmgray{1.757}
& 1.070\pmgray{0.540}
& 1.407\pmgray{0.379}
& 1.797\pmgray{1.058}
& \cellcolor{mygray}1.799\pmgray{1.256}\\
& Beards
& 0.375\pmgray{0.332}
& 1.175\pmgray{0.626}
& 1.735\pmgray{1.448}
& 1.108\pmgray{0.477}
& 1.837\pmgray{0.897}
& 2.167\pmgray{2.019}
& \cellcolor{mygray}1.952\pmgray{1.521}\\
& Blond Hair
& 0.457\pmgray{0.338}
& 1.010\pmgray{0.502}
& 1.324\pmgray{1.022}
& 0.736\pmgray{0.236}
& 0.916\pmgray{0.545}
& 1.524\pmgray{1.113}
& \cellcolor{mygray}1.528\pmgray{1.168}\\

\midrule
\multirow{5}{*}{VGG16}
& Smiling
& 0.891\pmgray{0.615}
& 1.454\pmgray{1.276}
& 3.742\pmgray{2.238}
& 1.240\pmgray{0.851}
& 1.459\pmgray{1.165}
& 4.684\pmgray{2.039}
& \cellcolor{mygray}3.628\pmgray{2.069}\\

& Big Nose
& 0.907\pmgray{0.248}
& 0.669\pmgray{0.609}
& 0.734\pmgray{0.464}
& 0.896\pmgray{0.721}
&  0.766\pmgray{0.528}
& 0.855\pmgray{0.583}
& \cellcolor{mygray}0.902\pmgray{0.482}\\

& Young
& 0.830\pmgray{0.747}
& 1.493\pmgray{0.856}
& 1.718\pmgray{0.991}
& 1.003\pmgray{0.760}
& 1.396\pmgray{0.721}
& 1.634\pmgray{0.916}
& \cellcolor{mygray}1.653\pmgray{1.067}\\

& Beards
& 0.873\pmgray{0.563}
& 1.193\pmgray{0.703}
& 1.729\pmgray{1.301}
& 1.234\pmgray{0.787}
& 2.095\pmgray{0.831}
& 2.222\pmgray{1.809}
& \cellcolor{mygray}2.101\pmgray{1.263}\\

& Blond Hair
& 0.251\pmgray{0.606}
& 0.923\pmgray{0.621}
& 1.132\pmgray{0.803}
& 0.619\pmgray{0.399}
& 0.920\pmgray{0.681}
& 1.092\pmgray{ 0.834}
& \cellcolor{mygray}1.461\pmgray{1.245}
\\

\midrule
\multirow{5}{*}{ViT-B/16}
& Smiling
& 1.144\pmgray{0.848}
& 1.301\pmgray{0.894}
& 2.366\pmgray{1.388}
& 1.640\pmgray{1.161}
& 1.404\pmgray{0.976}
& 2.426\pmgray{1.402}
& \cellcolor{mygray}2.371\pmgray{1.375}\\
& Big Nose
& 0.577\pmgray{0.419}
& 1.216\pmgray{1.093}
& 1.071\pmgray{0.704}
& 1.270\pmgray{0.796}
& 1.279\pmgray{0.943}
& 0.986\pmgray{0.693}
& \cellcolor{mygray}1.573\pmgray{1.039}\\
& Young
& 0.236\pmgray{0.135}
& 1.158\pmgray{0.841}
& 1.242\pmgray{0.735}
& 1.331\pmgray{1.181}
& 1.259\pmgray{0.974}
& 1.215\pmgray{0.939}
& \cellcolor{mygray}1.196\pmgray{0.997}\\
& Beards
& 0.248\pmgray{0.189}
& 1.351\pmgray{1.199}
& 1.383\pmgray{0.702}
& 1.171\pmgray{0.806}
& 1.589\pmgray{1.083}
& 1.548\pmgray{0.999}
& \cellcolor{mygray}1.176\pmgray{1.016}\\
& Blond Hair
& 0.763\pmgray{0.113}
& 0.943\pmgray{0.709}
& 1.138\pmgray{0.905}
& 0.753\pmgray{0.511}
& 1.069\pmgray{1.031}
& 0.908\pmgray{0.680}
& \cellcolor{mygray}1.061\pmgray{0.717}\\
   
\bottomrule
\end{tabular}
}
\caption{
Editing Precision Ratio (EPR) scores for different methods and attribute-classifier combinations.
}
\label{tab:epr_full}
\end{table*}

\begin{table*}[ht]
\centering
\renewcommand{\arraystretch}{1.2}
\definecolor{mygray}{gray}{0.92}
\definecolor{lightgray}{gray}{0.4}
\newcommand{\pmgray}[1]{\textcolor{lightgray}{\text{\scriptsize$\pm$#1}}}
\resizebox{\textwidth}{!}{
\begin{tabular}{l l |c|ccccc|c}
\toprule
\multirow{2}{*}{\textbf{Concept}} &
\multirow{2}{*}{\textbf{Metric}} &
\multirow{2}{*}{\textbf{Recon.}} &
\multicolumn{5}{c|}{\textbf{Editing Methods}} &
\multicolumn{1}{c}{\textbf{Explainable}} \\
\cmidrule(lr){4-8} \cmidrule(lr){9-9}
& & &
\textbf{Boundary} &
\textbf{Asyrp} &
\textbf{Slider} &
\textbf{MasaCtrl} &
\textbf{SwiftEdit} &
\textbf{CASL-Steer} \\

\midrule
\multirow{2}{*}{Gothic}
& CLIP-Score~($\uparrow$)
& 0.103\pmgray{0.014}
& 0.129\pmgray{0.021}
& 0.158\pmgray{0.118}
& 0.167\pmgray{0.014}
& 0.193\pmgray{0.039}
& 0.184\pmgray{0.091}
& 0.172\pmgray{0.031} \\   
& LPIPS~($\downarrow$)
& 0.184\pmgray{0.028}
& 0.268\pmgray{0.052}
& 0.302\pmgray{0.143}
& 0.256\pmgray{0.035}
& 0.332\pmgray{0.071}
& 0.315\pmgray{0.152}
& 0.247\pmgray{0.040} \\

\midrule
\multirow{2}{*}{Red Brick}
& CLIP-Score~($\uparrow$)
& 0.118\pmgray{0.016}
& 0.145\pmgray{0.024}
& 0.174\pmgray{0.132}
& 0.183\pmgray{0.016}
& 0.196\pmgray{0.042}
& 0.211\pmgray{0.097}
& 0.203\pmgray{0.037} \\  
& LPIPS~($\downarrow$)
& 0.198\pmgray{0.030}
& 0.279\pmgray{0.060}
& 0.310\pmgray{0.149}
& 0.276\pmgray{0.038}
& 0.338\pmgray{0.078}
& 0.327\pmgray{0.162}
& 0.251\pmgray{0.044} \\

\midrule
\multirow{2}{*}{Snowy}
& CLIP-Score~($\uparrow$)
& 0.132\pmgray{0.019}
& 0.154\pmgray{0.027}
& 0.182\pmgray{0.146}
& 0.189\pmgray{0.018}
& 0.219\pmgray{0.049}
& 0.226\pmgray{0.124}
& 0.209\pmgray{0.034} \\  
& LPIPS~($\downarrow$)
& 0.213\pmgray{0.033}
& 0.291\pmgray{0.067}
& 0.326\pmgray{0.164}
& 0.284\pmgray{0.041}
& 0.353\pmgray{0.083}
& 0.341\pmgray{0.174}
& 0.258\pmgray{0.048} \\

\midrule
\multirow{2}{*}{Wooden}
& CLIP-Score~($\uparrow$)
& 0.111\pmgray{0.015}
& 0.139\pmgray{0.023}
& 0.164\pmgray{0.128}
& 0.174\pmgray{0.017}
& 0.204\pmgray{0.047}
& 0.198\pmgray{0.102}
& 0.218\pmgray{0.052} \\   
& LPIPS~($\downarrow$)
& 0.191\pmgray{0.031}
& 0.273\pmgray{0.058}
& 0.298\pmgray{0.156}
& 0.268\pmgray{0.036}
& 0.329\pmgray{0.080}
& 0.316\pmgray{0.170}
& 0.244\pmgray{0.043} \\

\bottomrule
\end{tabular}
}
\caption{Evaluation on LSUN-Church dataset.}
\label{tab:statistic_result_all}
\end{table*}

\begin{table*}[ht]
\centering
\renewcommand{\arraystretch}{1.2}
\definecolor{mygray}{gray}{0.92}
\definecolor{lightgray}{gray}{0.4}
\newcommand{\pmgray}[1]{\textcolor{lightgray}{\text{\scriptsize$\pm$#1}}}
\resizebox{\textwidth}{!}{
\begin{tabular}{l l |c|ccccc|c}
\toprule
\multirow{2}{*}{\textbf{Concept}} &
\multirow{2}{*}{\textbf{Metric}} &
\multirow{2}{*}{\textbf{Recon.}} &
\multicolumn{5}{c|}{\textbf{Editing Methods}} &
\multicolumn{1}{c}{\textbf{Explainable}} \\
\cmidrule(lr){4-8} \cmidrule(lr){9-9}
& & &
\textbf{Boundary} &
\textbf{Asyrp} &
\textbf{Slider} &
\textbf{MasaCtrl} &
\textbf{SwiftEdit} &
\textbf{CASL-Steer} \\

\midrule
\multirow{2}{*}{Angry}
& CLIP-Score~($\uparrow$)
& 0.115\pmgray{0.017}
& 0.142\pmgray{0.024}
& 0.171\pmgray{0.148}
& 0.182\pmgray{0.016}
& 0.201\pmgray{0.045}
& 0.219\pmgray{0.134}
& 0.197\pmgray{0.039} \\   
& LPIPS~($\downarrow$)
& 0.204\pmgray{0.031}
& 0.284\pmgray{0.062}
& 0.318\pmgray{0.173}
& 0.276\pmgray{0.039}
& 0.347\pmgray{0.087}
& 0.333\pmgray{0.182}
& 0.253\pmgray{0.046} \\

\midrule
\multirow{2}{*}{Sleepy}
& CLIP-Score~($\uparrow$)
& 0.108\pmgray{0.015}
& 0.131\pmgray{0.020}
& 0.159\pmgray{0.136}
& 0.167\pmgray{0.014}
& 0.185\pmgray{0.041}
& 0.193\pmgray{0.121}
& 0.176\pmgray{0.033} \\   
& LPIPS~($\downarrow$)
& 0.196\pmgray{0.030}
& 0.273\pmgray{0.057}
& 0.301\pmgray{0.159}
& 0.268\pmgray{0.036}
& 0.329\pmgray{0.082}
& 0.317\pmgray{0.168}
& 0.243\pmgray{0.045} \\

\midrule
\multirow{2}{*}{Smiling}
& CLIP-Score~($\uparrow$)
& 0.121\pmgray{0.018}
& 0.148\pmgray{0.025}
& 0.179\pmgray{0.155}
& 0.188\pmgray{0.017}
& 0.208\pmgray{0.047}
& 0.217\pmgray{0.145}
& 0.204\pmgray{0.041} \\  
& LPIPS~($\downarrow$)
& 0.207\pmgray{0.033}
& 0.289\pmgray{0.064}
& 0.324\pmgray{0.166}
& 0.277\pmgray{0.040}
& 0.346\pmgray{0.089}
& 0.335\pmgray{0.178}
& 0.257\pmgray{0.048} \\

\midrule
\multirow{2}{*}{Puppy-like}
& CLIP-Score~($\uparrow$)
& 0.134\pmgray{0.019}
& 0.156\pmgray{0.027}
& 0.188\pmgray{0.162}
& 0.197\pmgray{0.018}
& 0.221\pmgray{0.052}
& 0.226\pmgray{0.152}
& 0.233\pmgray{0.056} \\   
& LPIPS~($\downarrow$)
& 0.215\pmgray{0.034}
& 0.296\pmgray{0.069}
& 0.332\pmgray{0.178}
& 0.283\pmgray{0.041}
& 0.359\pmgray{0.093}
& 0.341\pmgray{0.185}
& 0.249\pmgray{0.050} \\

\bottomrule
\end{tabular}
}
\caption{Evaluation on the AFHQ-Dog dataset.}
\label{tab:statistic_result_dog}
\end{table*}

\section{Full Hyperparameter Analysis Results}

Due to space limitations in the main text, we provide the complete results of our hyperparameter analysis in the supplementary material.

\begin{figure}[ht]
    \centering
    \includegraphics[width=0.98\linewidth]{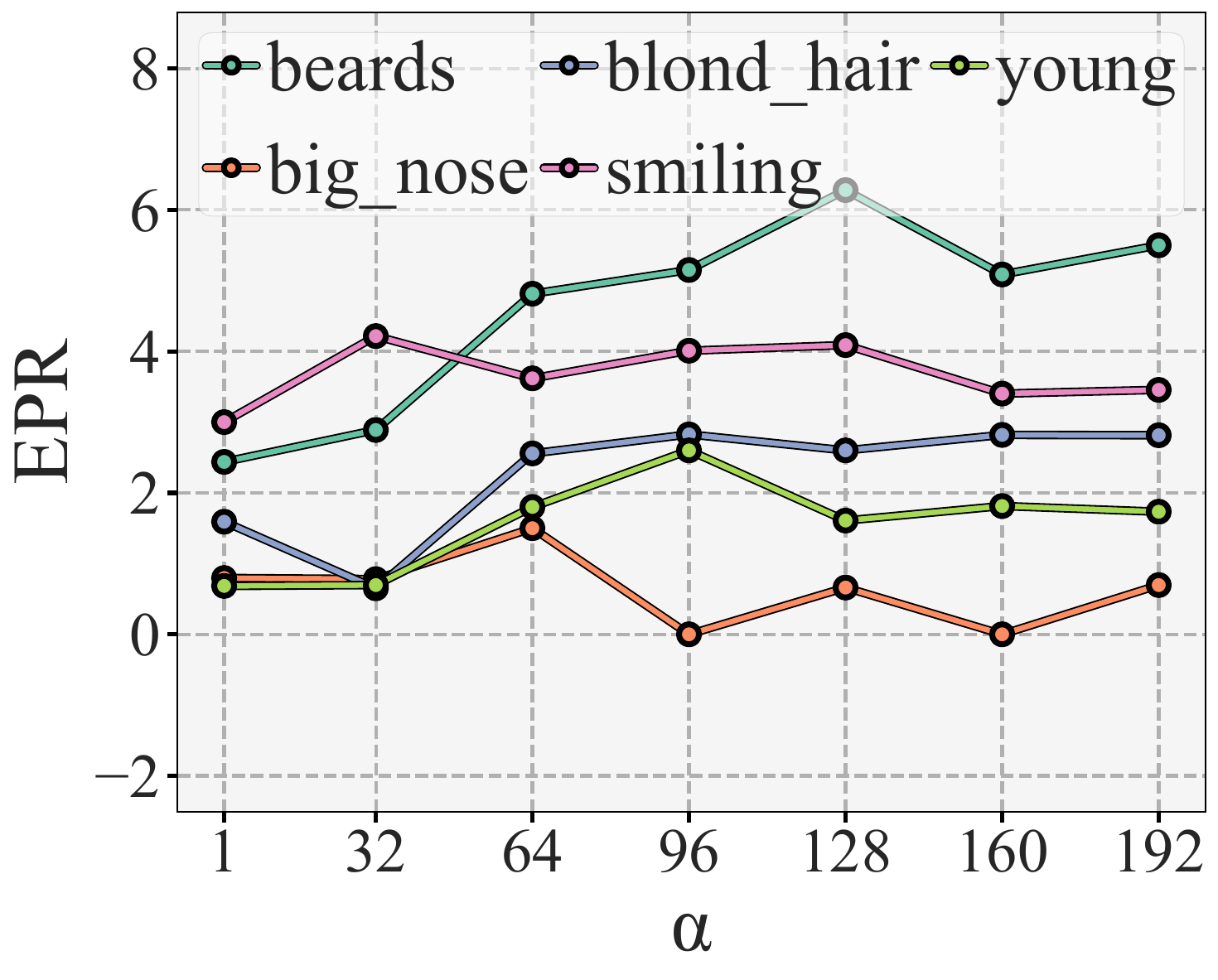}
    \caption{EPR vs. $\alpha$ for top-$k=1$ (all concepts).}
\end{figure}

\begin{figure}[ht]
    \centering
    \includegraphics[width=0.98\linewidth]{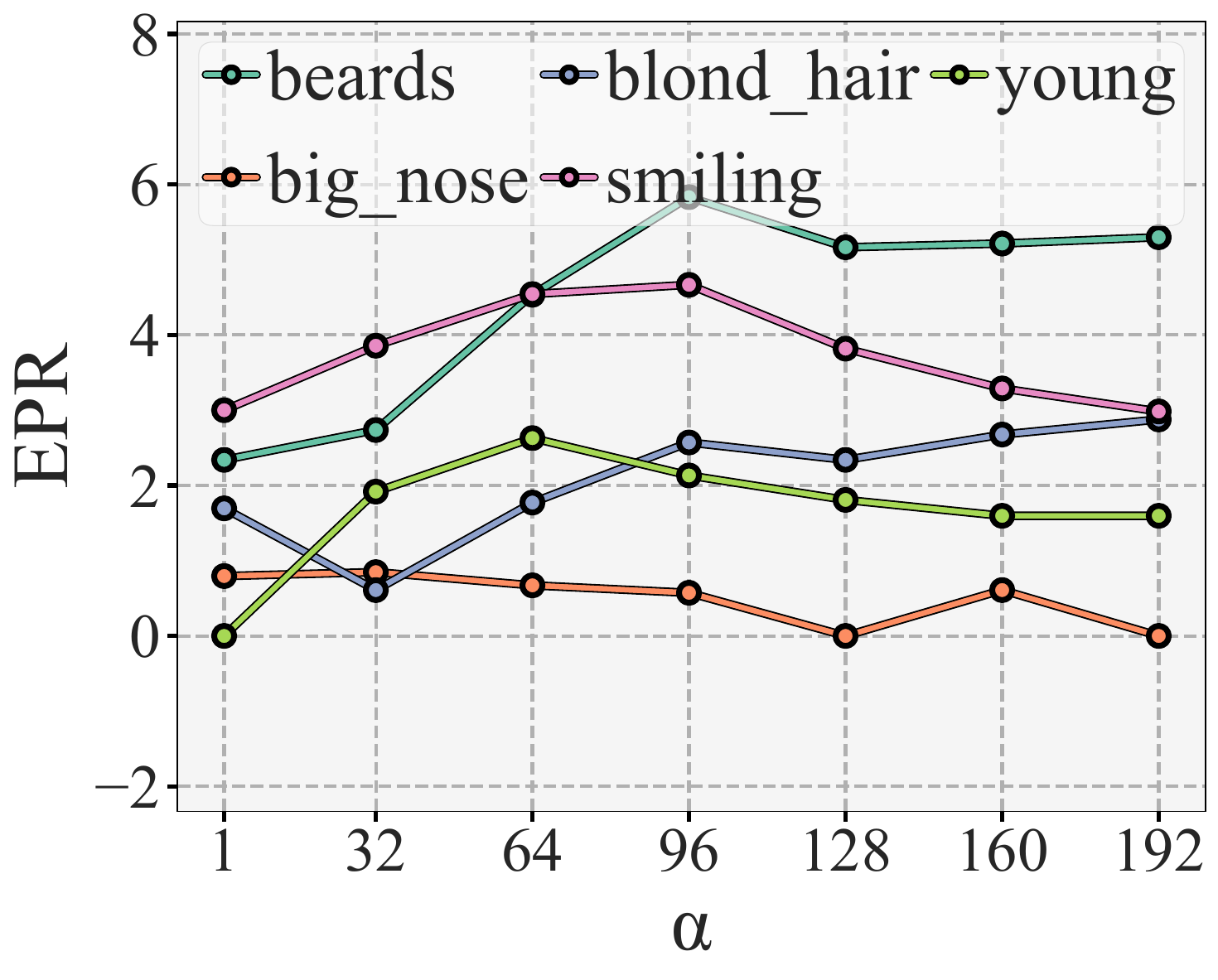}
    \caption{EPR vs. $\alpha$ for top-$k=2$ (all concepts).}
\end{figure}

\begin{figure}[ht]
    \centering
    \includegraphics[width=0.98\linewidth]{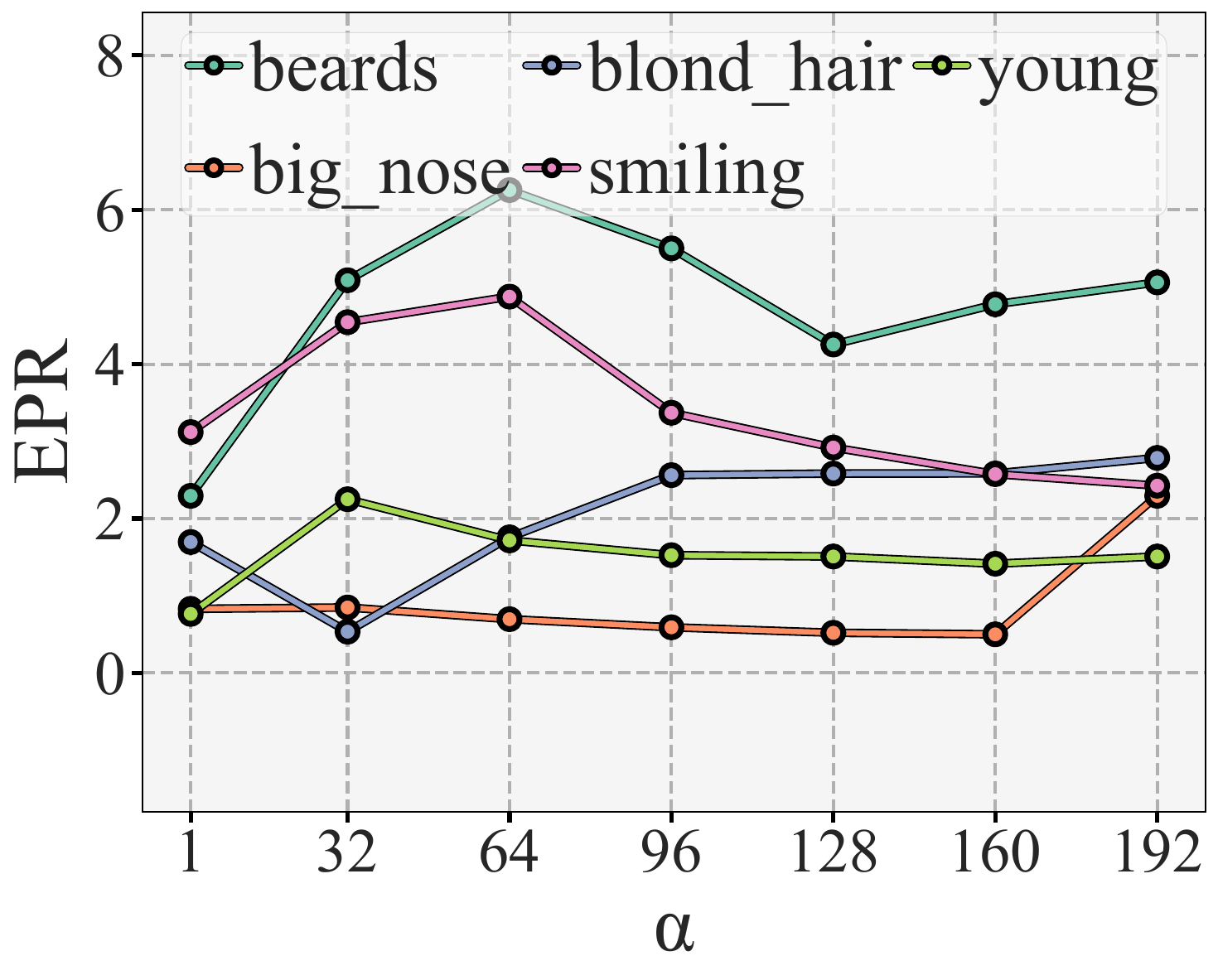}
    \caption{EPR vs. $\alpha$ for top-$k=4$ (all concepts).}
\end{figure}

\begin{figure}[ht]
    \centering
    \includegraphics[width=0.98\linewidth]{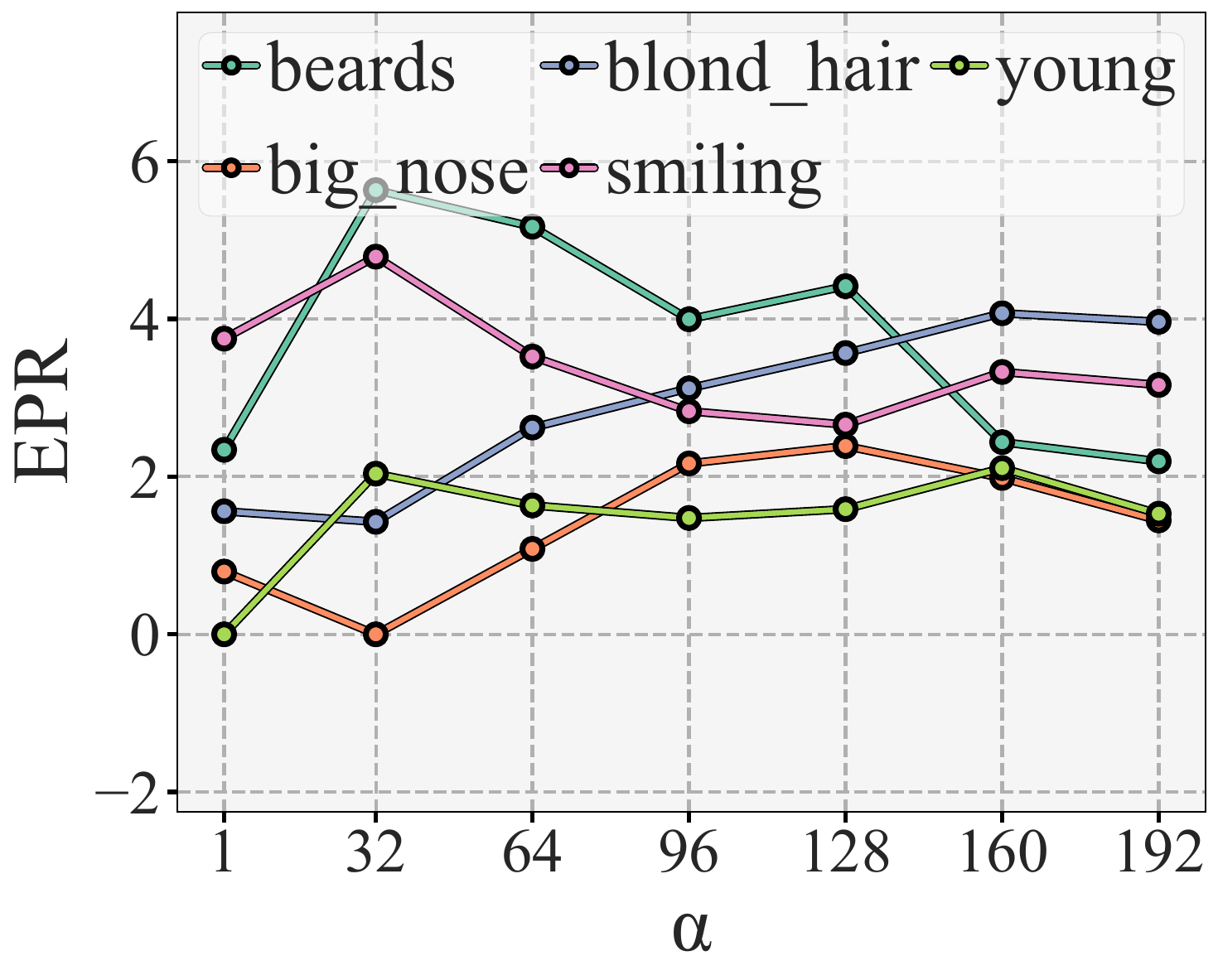}
    \caption{EPR vs. $\alpha$ for top-$k=8$ (all concepts).}
\end{figure}

\begin{figure}[ht]
    \centering
    \includegraphics[width=0.98\linewidth]{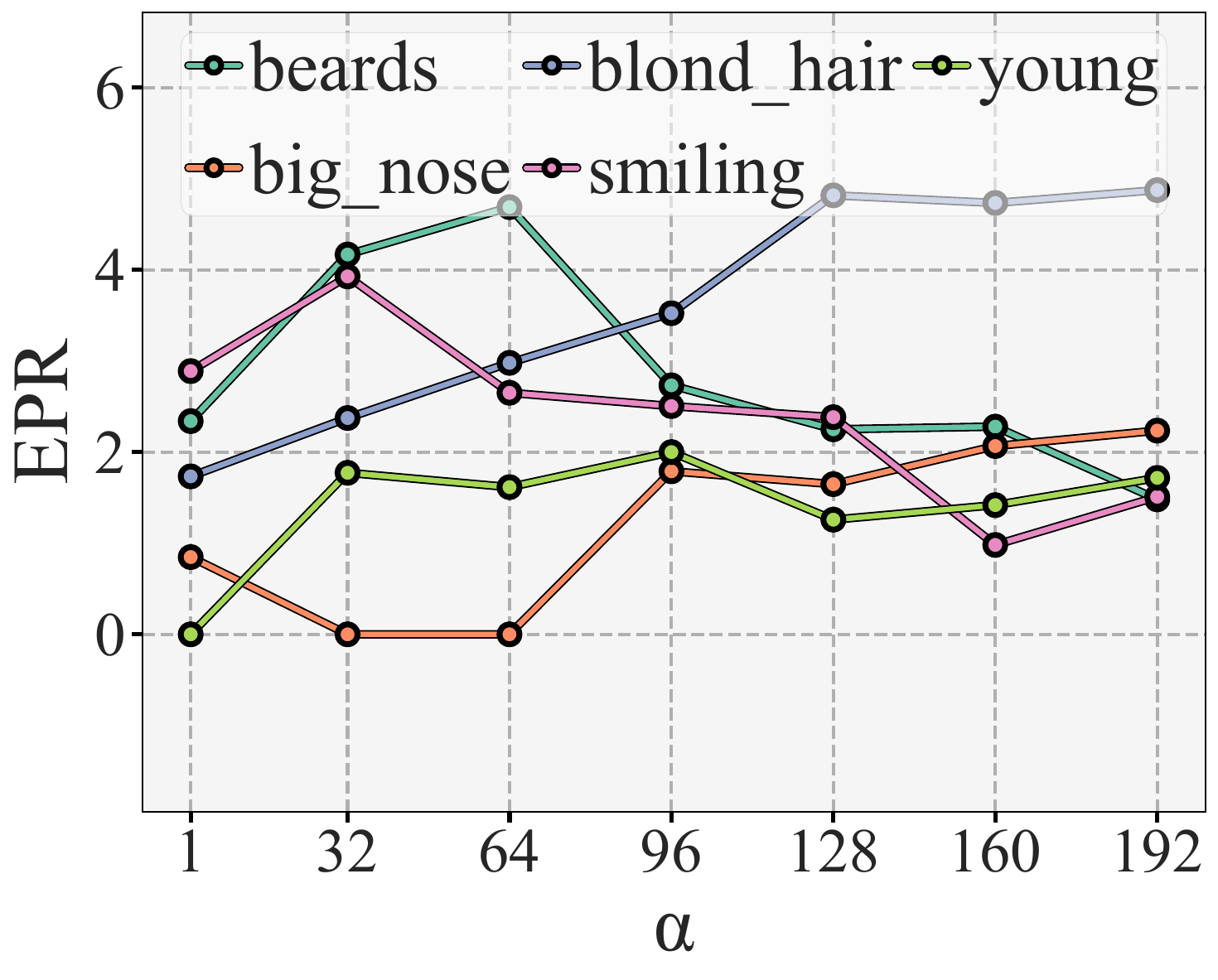}
    \caption{EPR vs. $\alpha$ for top-$k=16$ (all concepts).}
\end{figure}

\begin{figure}[ht]
    \centering
    \includegraphics[width=0.98\linewidth]{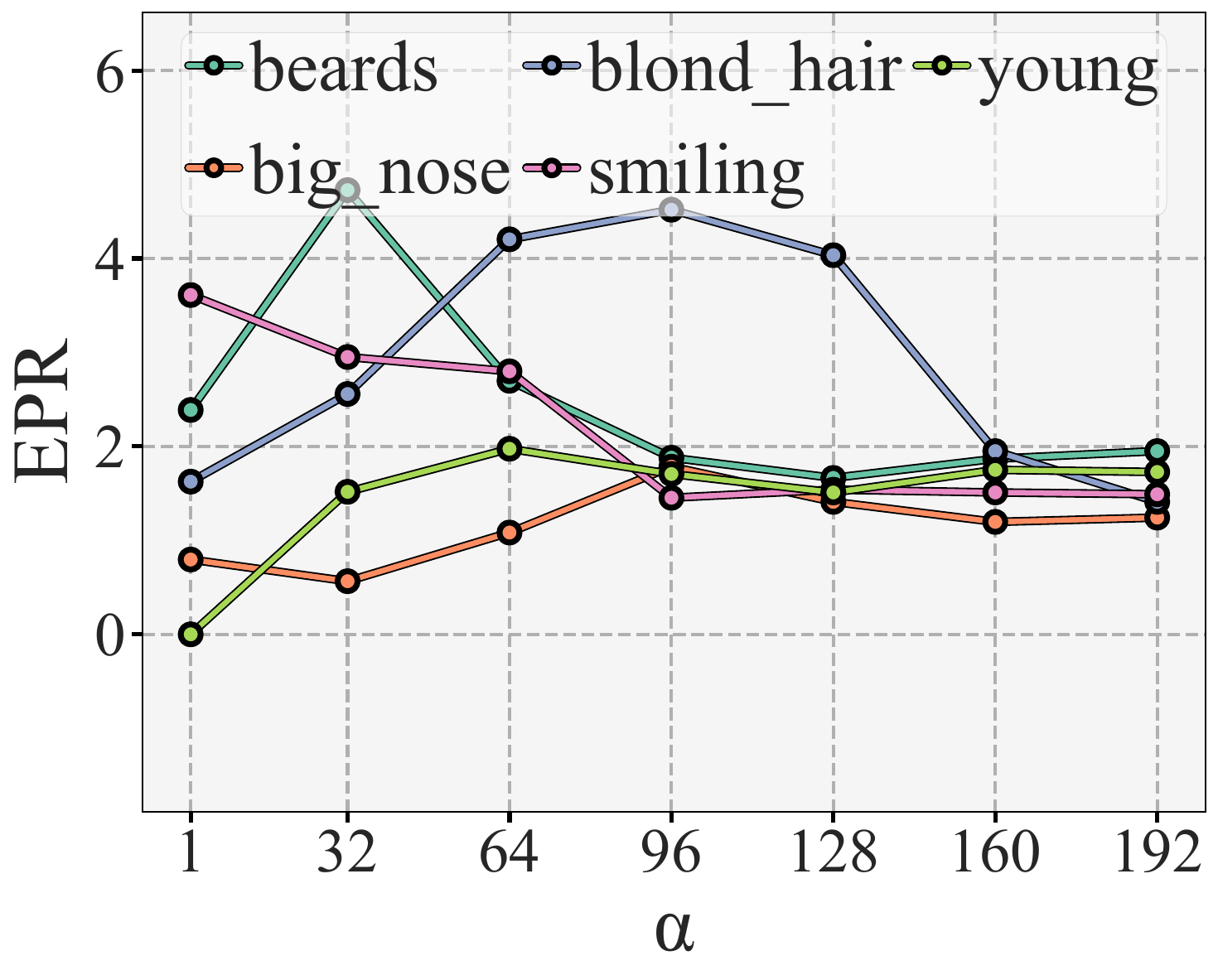}
    \caption{EPR vs. $\alpha$ for top-$k=32$ (all concepts).}
\end{figure}

\begin{figure}[ht]
    \centering
    \includegraphics[width=0.98\linewidth]{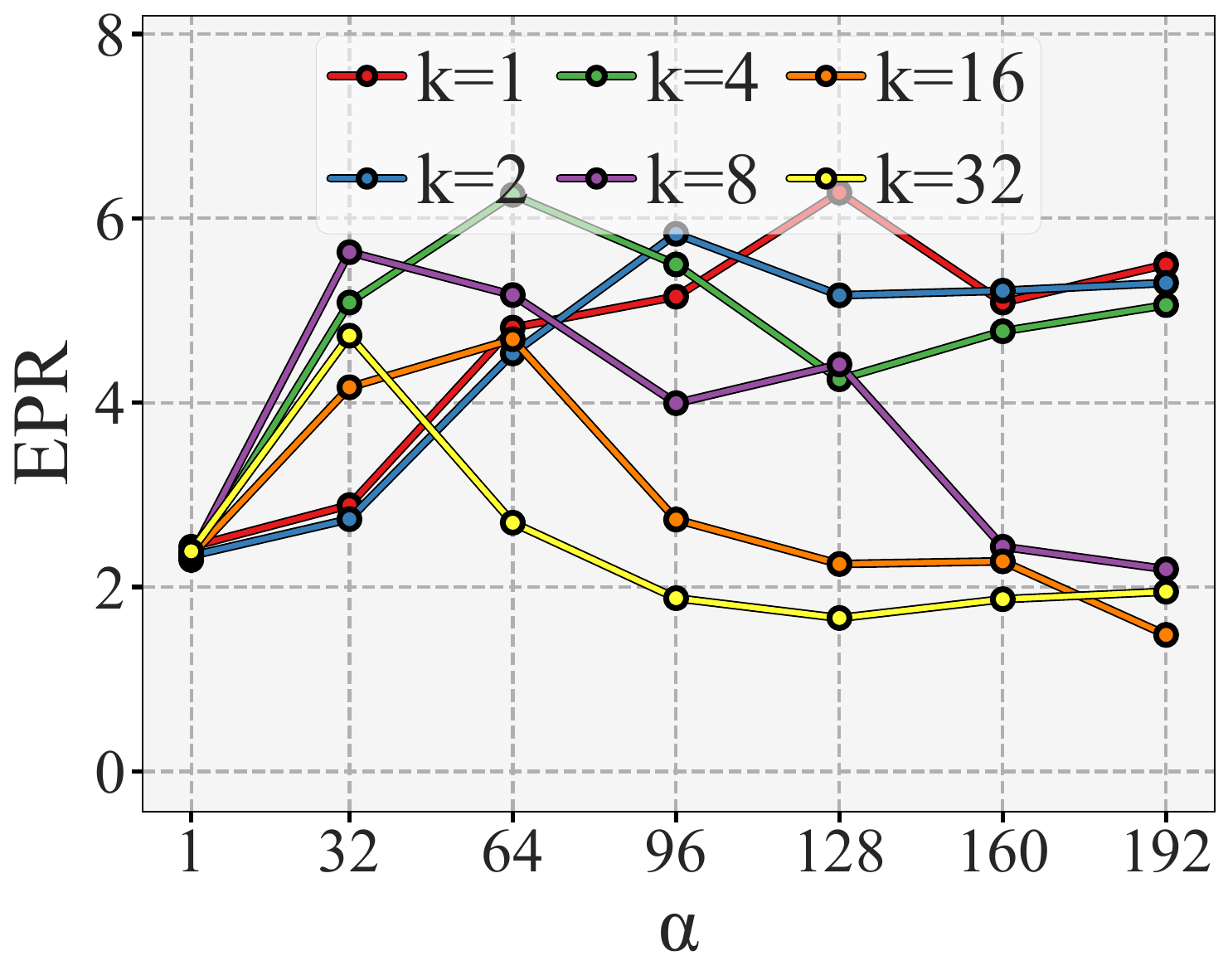}
    \caption{EPR vs. $\alpha$ for Beards.}
\end{figure}

\begin{figure}[ht]
    \centering
    \includegraphics[width=0.98\linewidth]{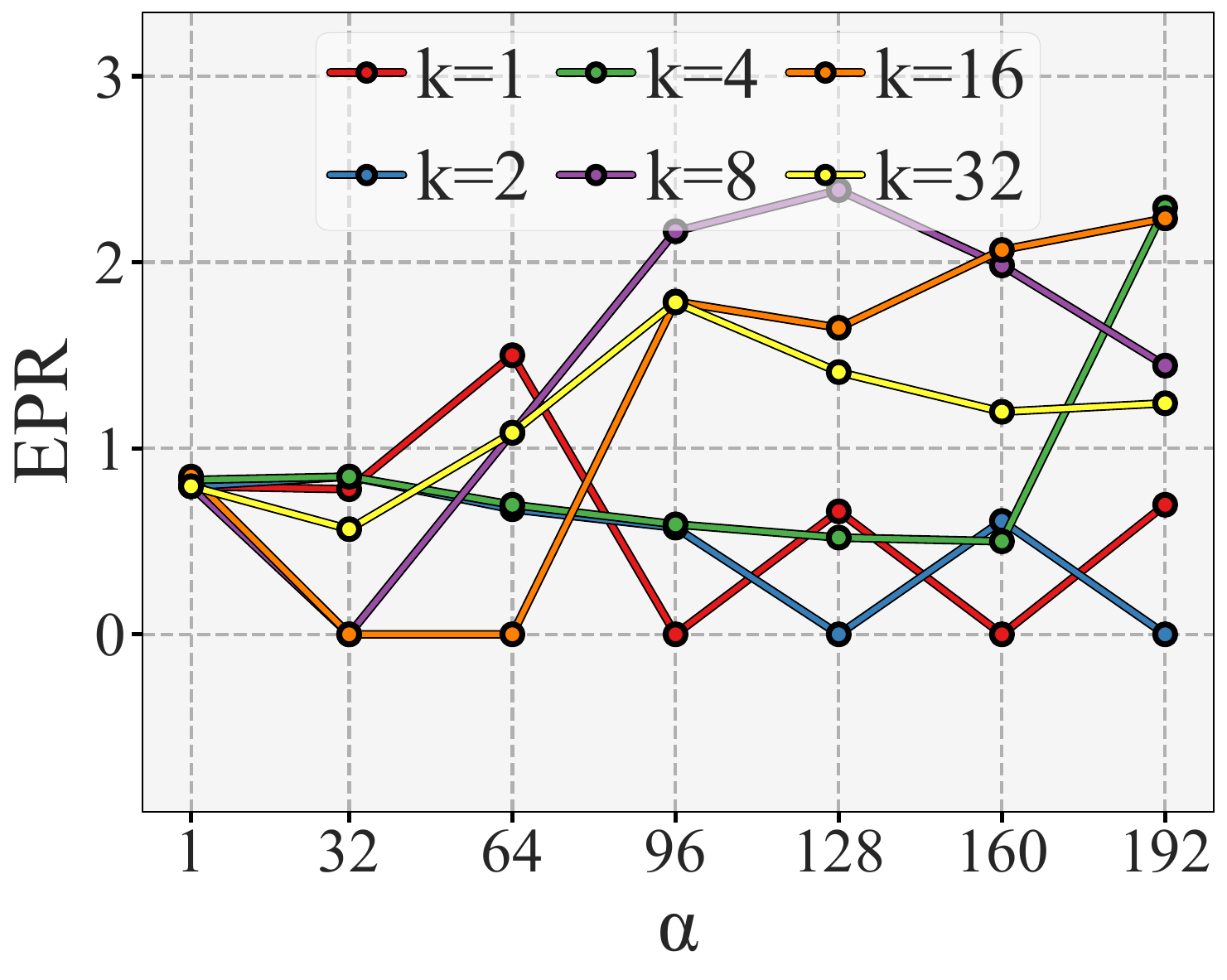}
    \caption{EPR vs. $\alpha$ for Big Nose.}
\end{figure}

\begin{figure}[ht]
    \centering
    \includegraphics[width=0.98\linewidth]{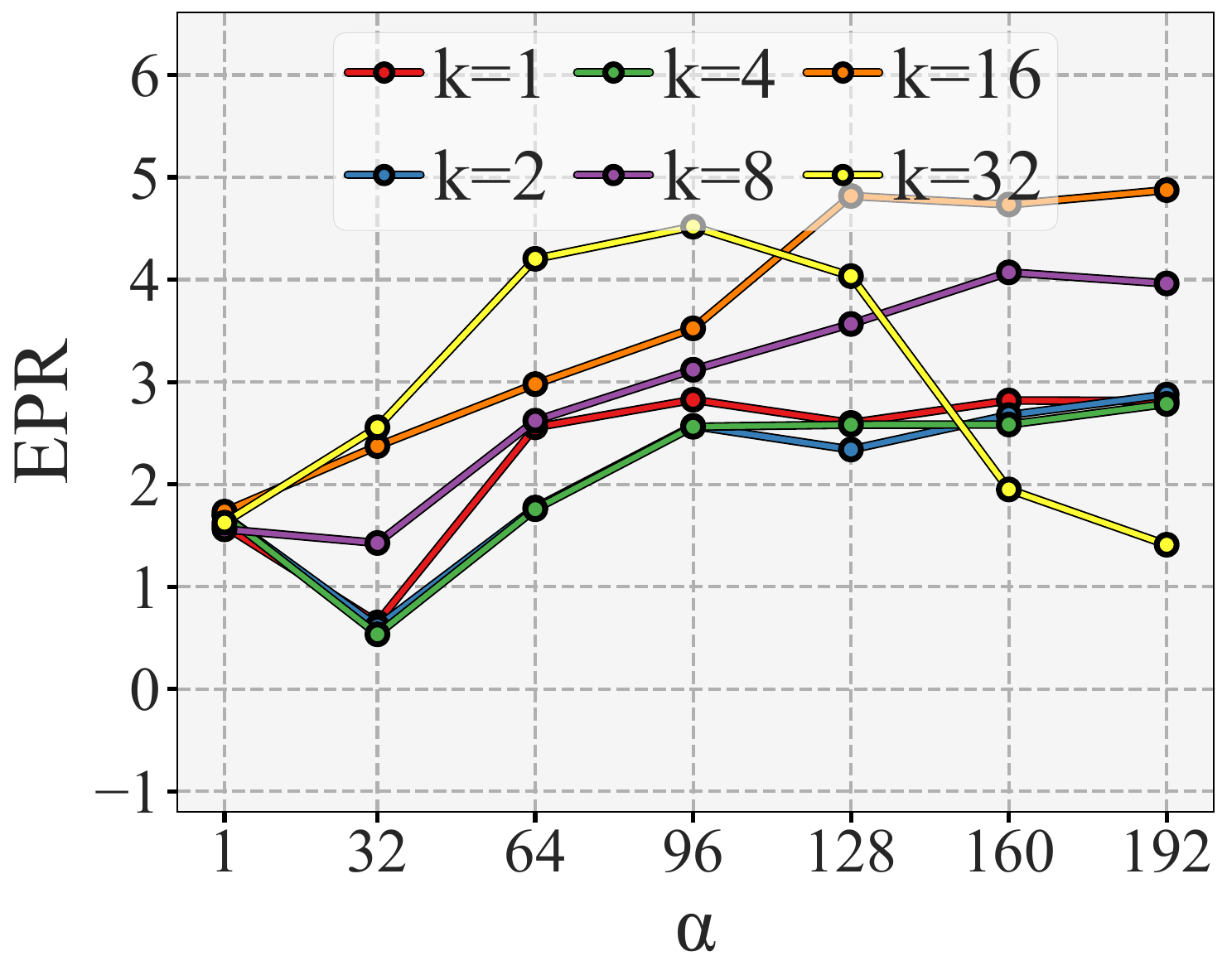}
    \caption{EPR vs. $\alpha$ for Blond Hair.}
\end{figure}

\begin{figure}[ht]
    \centering
    \includegraphics[width=0.98\linewidth]{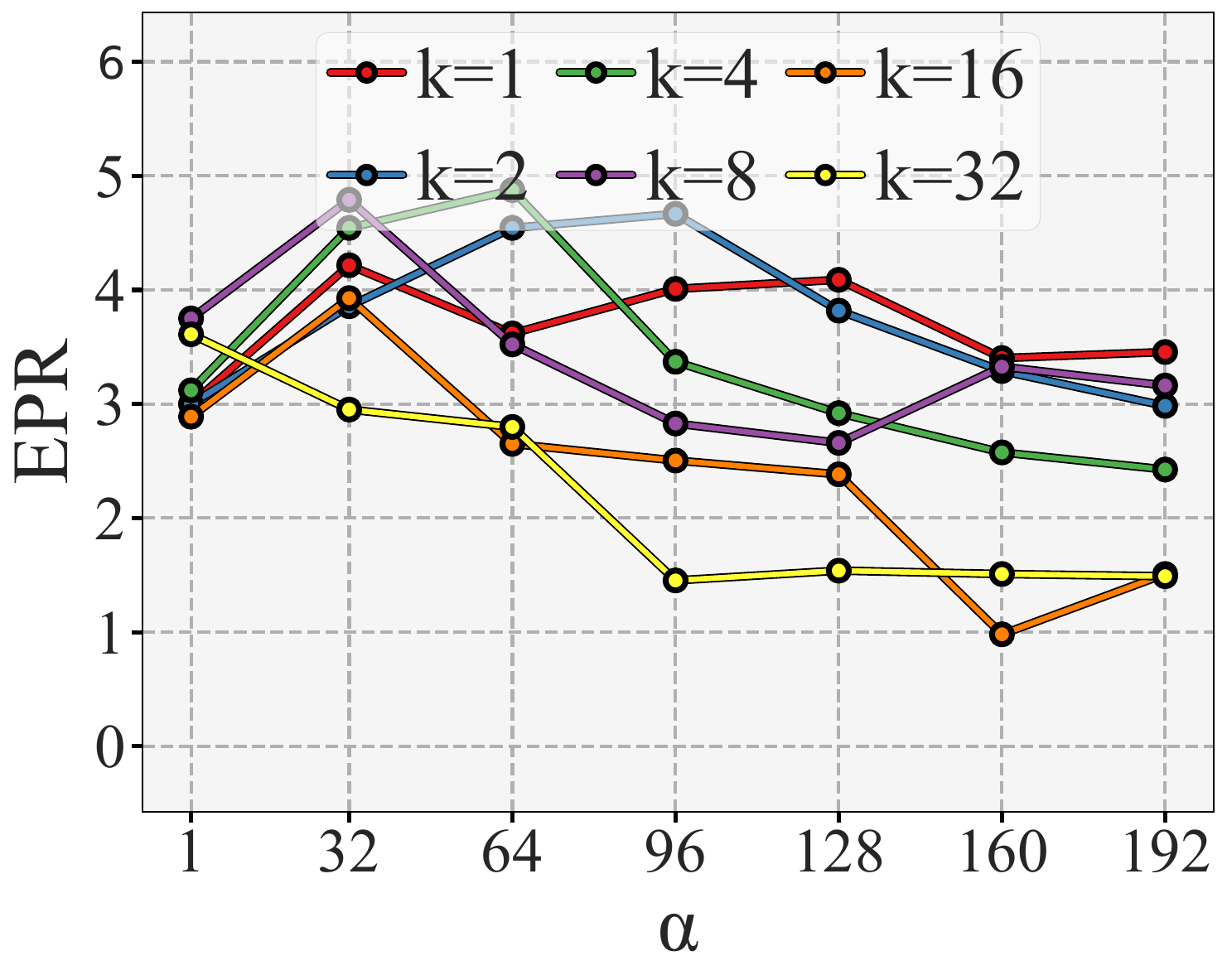}
    \caption{EPR vs. $\alpha$ for Smiling.}
\end{figure}

\begin{figure}[ht]
    \centering
    \includegraphics[width=0.98\linewidth]{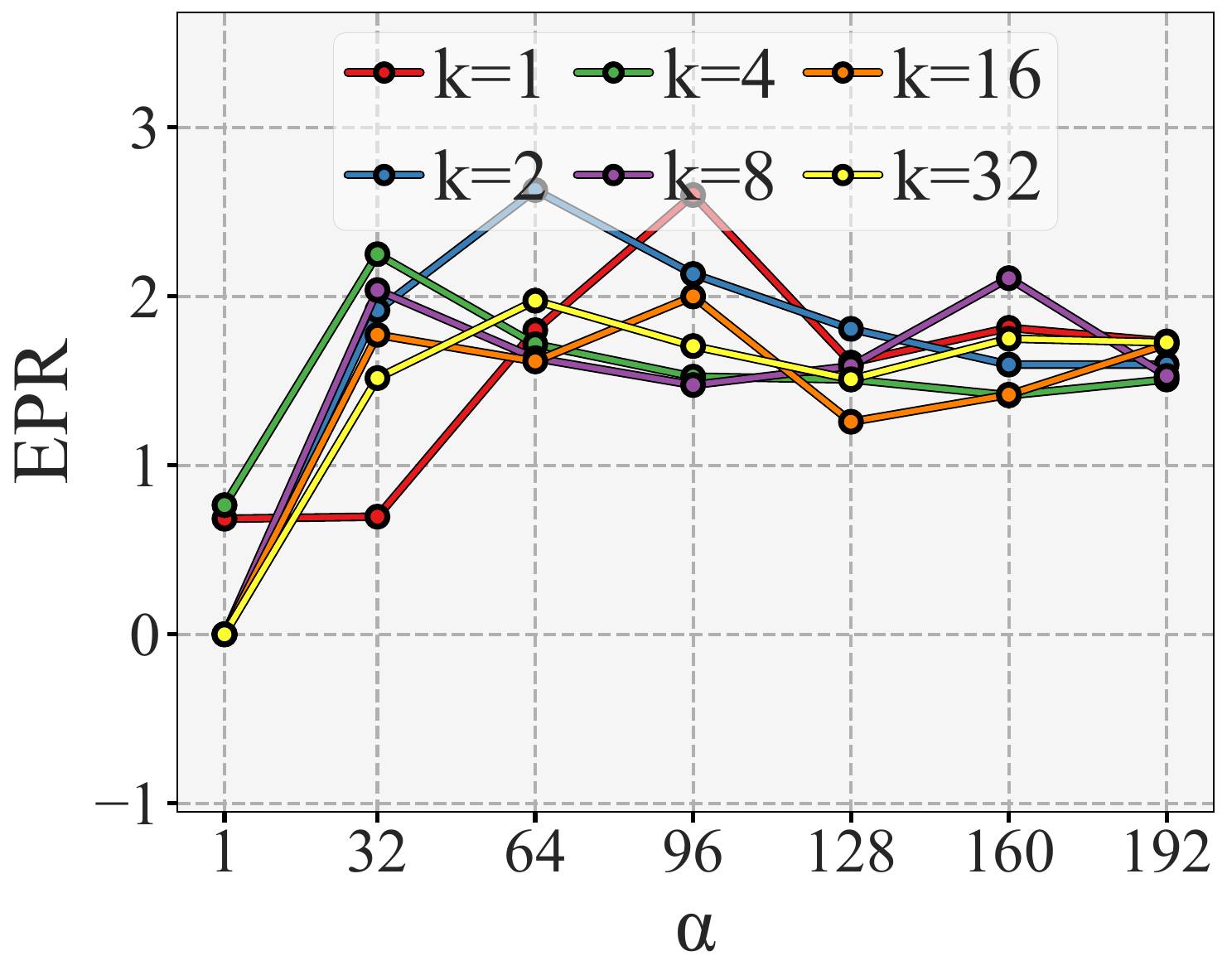}
    \caption{EPR vs. $\alpha$ for Young.}
\end{figure}

\begin{figure}[ht]
    \centering
    \includegraphics[width=0.98\linewidth]{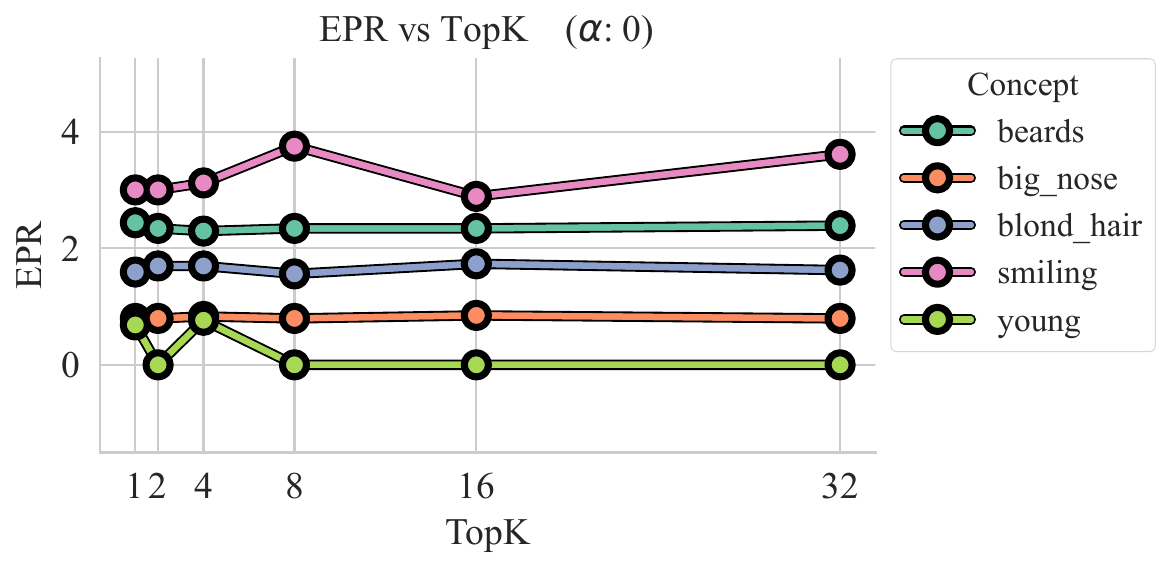}
    \caption{EPR vs. top-$k$ for $\alpha=0$.}
\end{figure}

\begin{figure}[ht]
    \centering
    \includegraphics[width=0.98\linewidth]{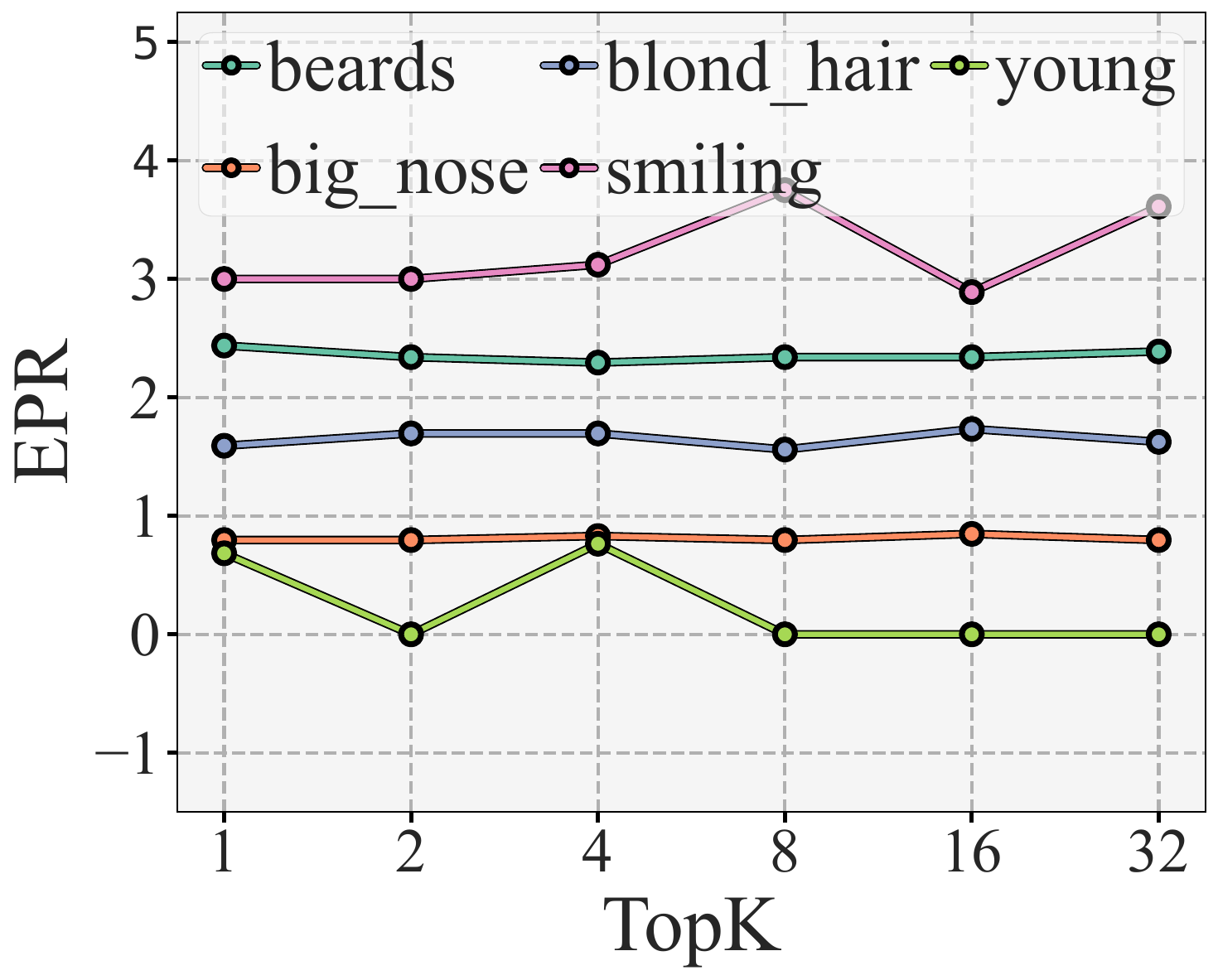}
    \caption{EPR vs. top-$k$ for $\alpha=1$.}
\end{figure}

\begin{figure}[ht]
    \centering
    \includegraphics[width=0.98\linewidth]{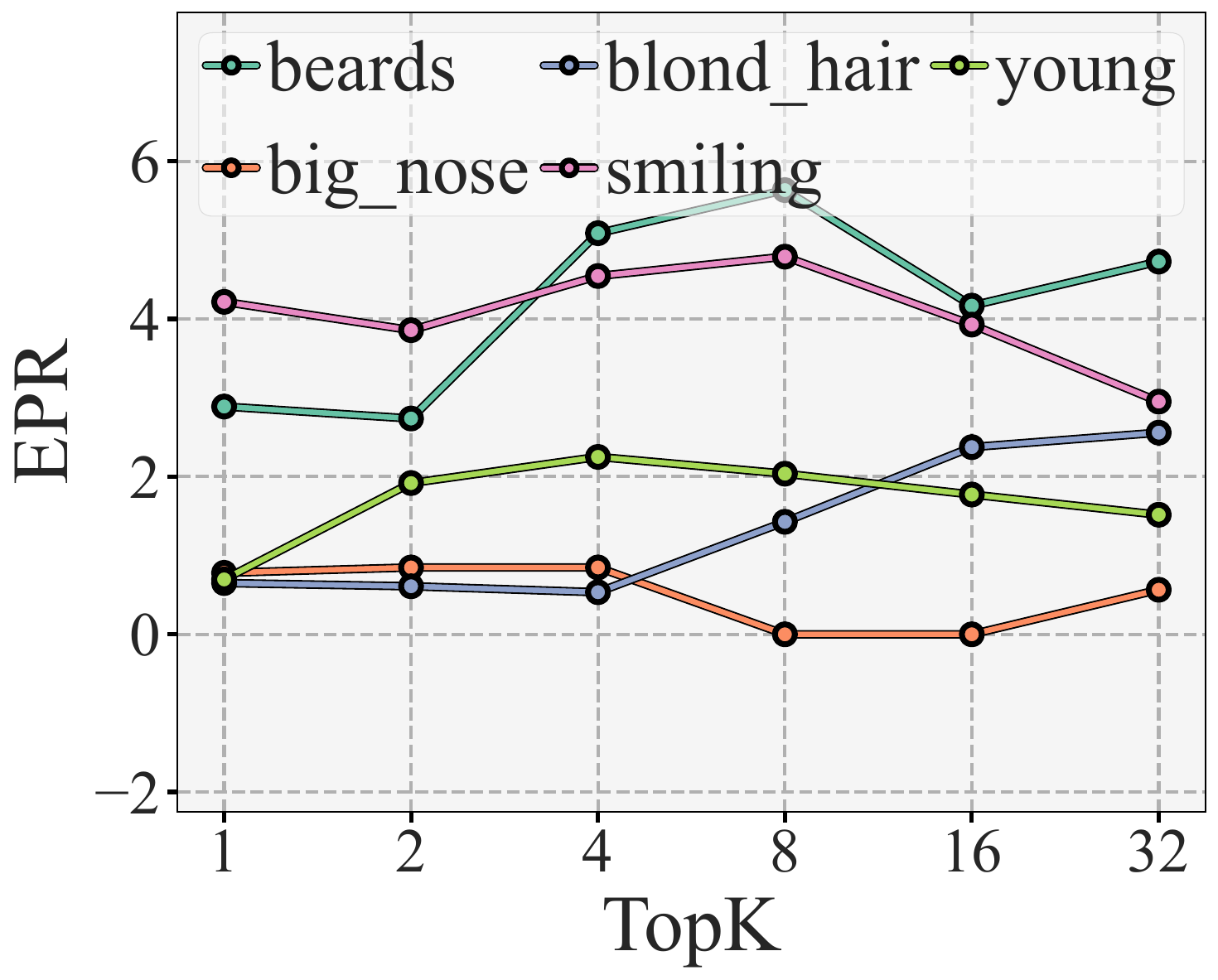}
    \caption{EPR vs. top-$k$ for $\alpha=32$.}
\end{figure}

\begin{figure}[ht]
    \centering
    \includegraphics[width=0.98\linewidth]{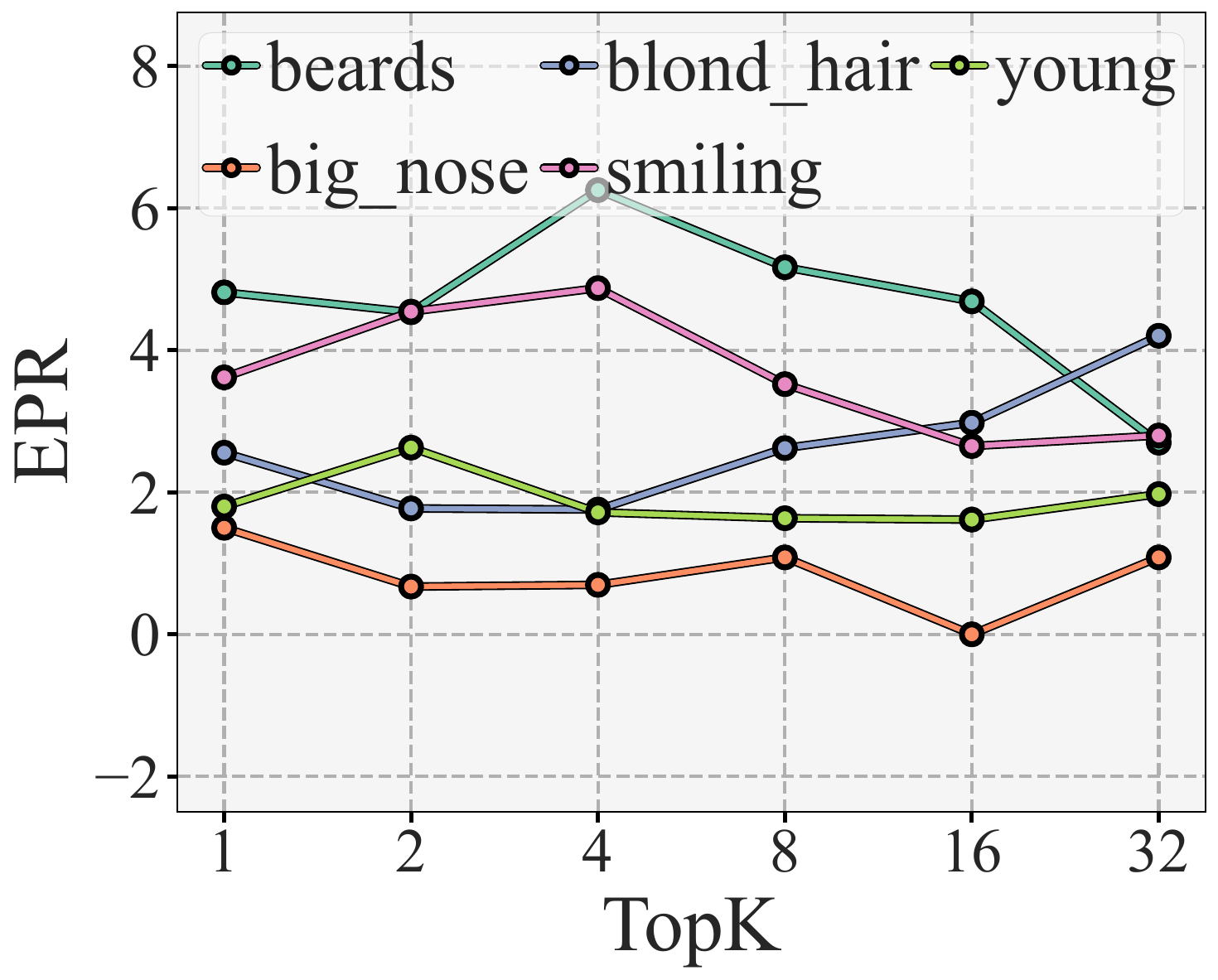}
    \caption{EPR vs. top-$k$ for $\alpha=64$.}
\end{figure}

\begin{figure}[ht]
    \centering
    \includegraphics[width=0.98\linewidth]{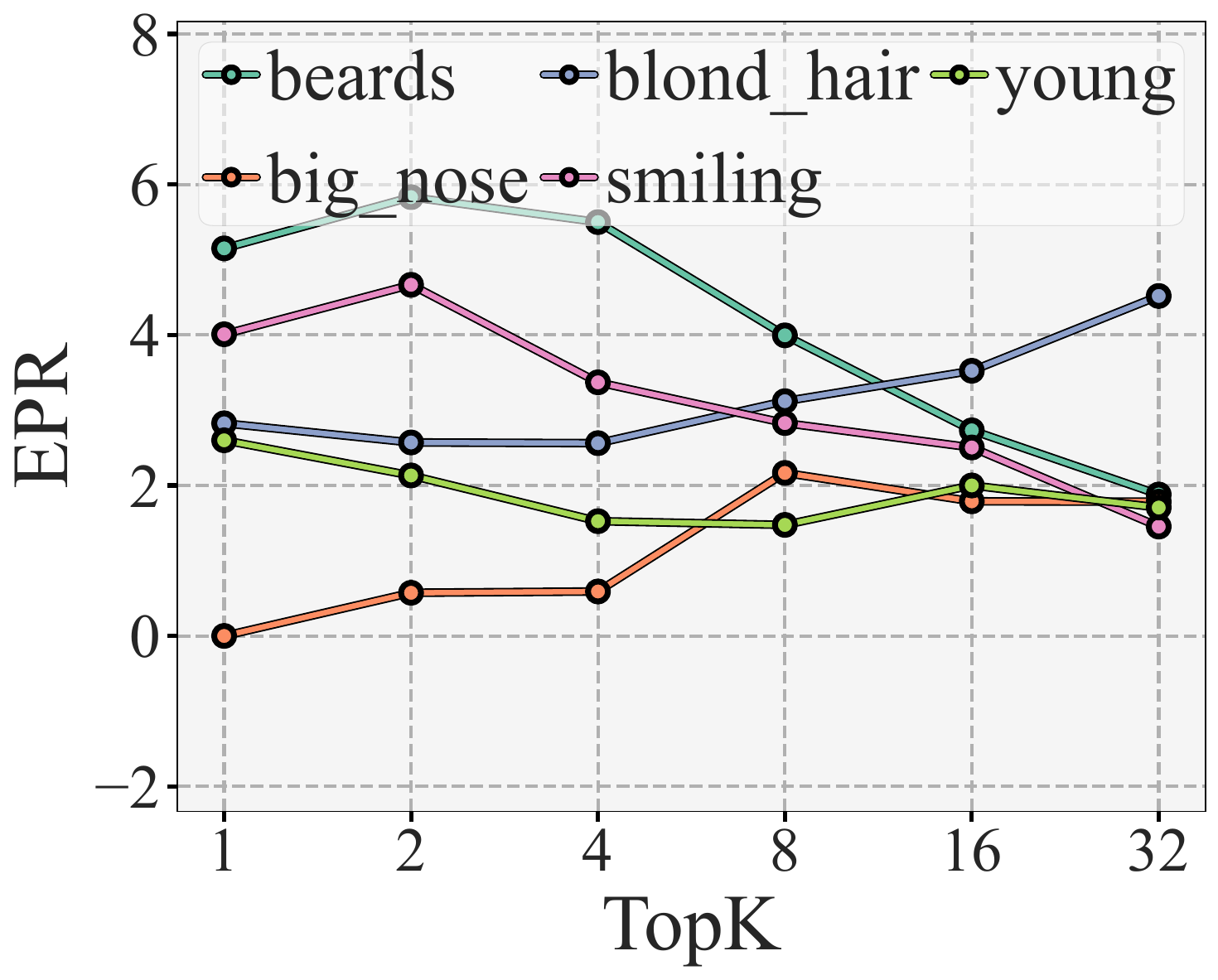}
    \caption{EPR vs. top-$k$ for $\alpha=96$.}
\end{figure}

\begin{figure}[ht]
    \centering
    \includegraphics[width=0.98\linewidth]{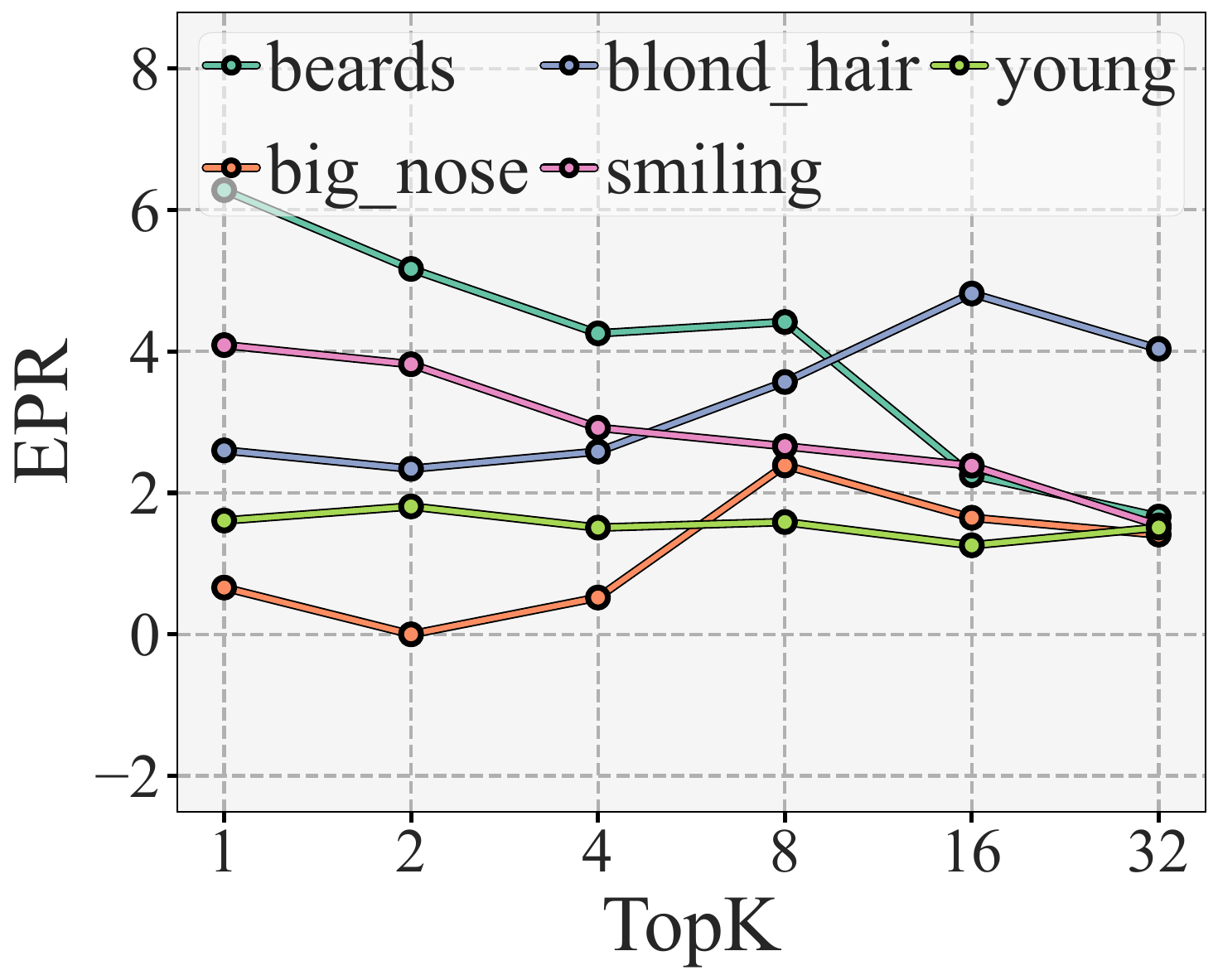}
    \caption{EPR vs. top-$k$ for $\alpha=128$.}
\end{figure}

\begin{figure}[ht]
    \centering
    \includegraphics[width=0.98\linewidth]{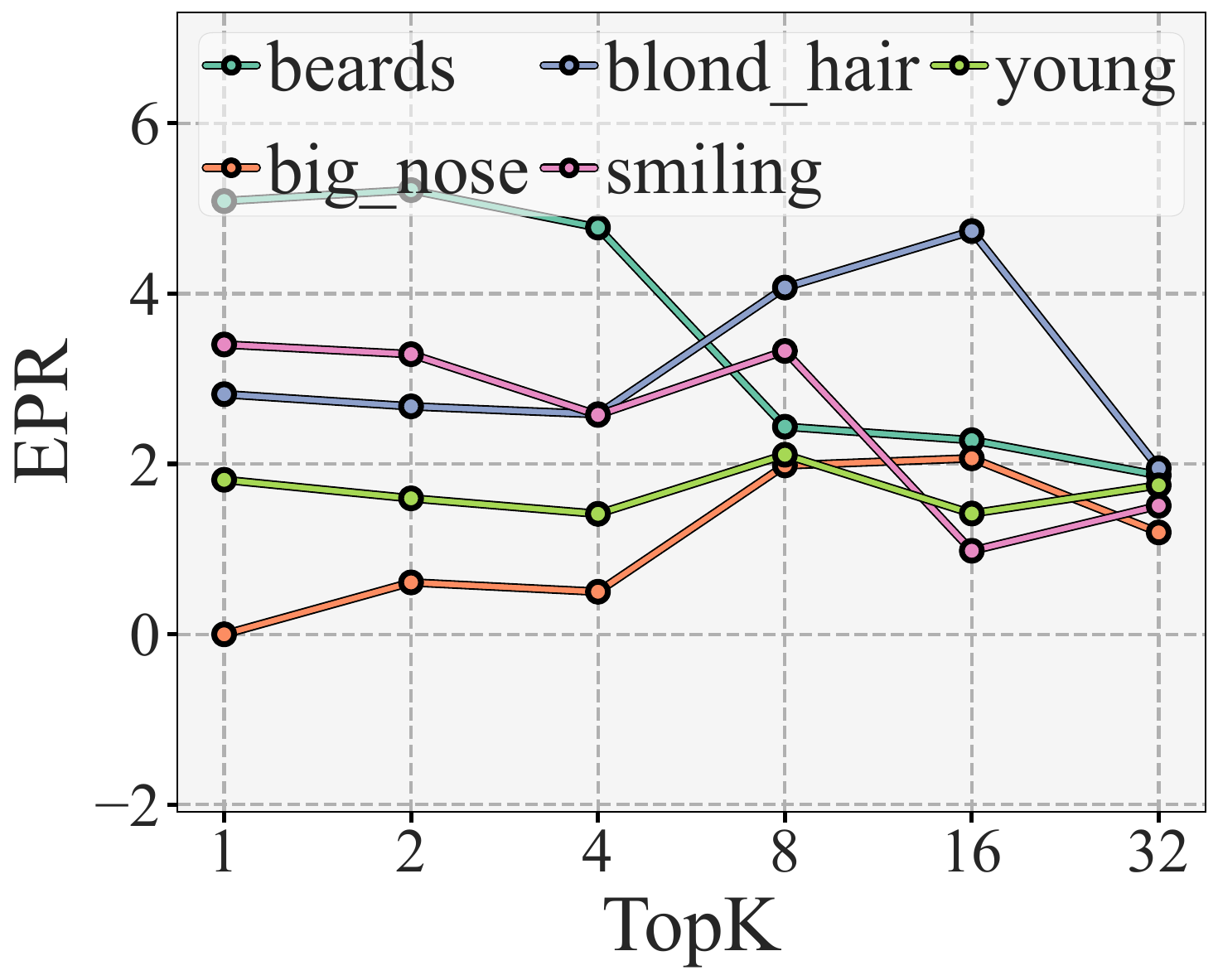}
    \caption{EPR vs. top-$k$ for $\alpha=160$.}
\end{figure}

\begin{figure}[ht]
    \centering
    \includegraphics[width=0.98\linewidth]{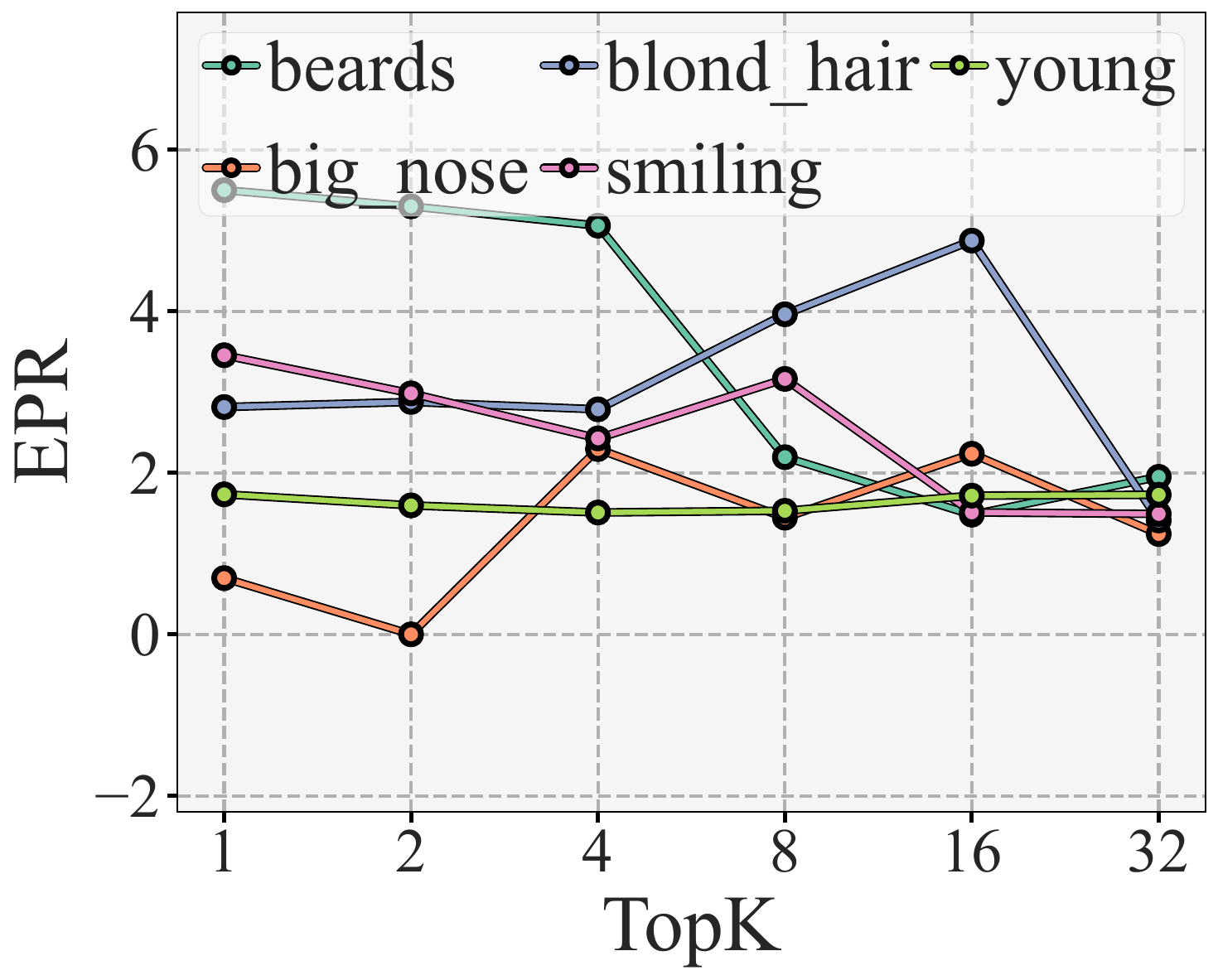}
    \caption{EPR vs. top-$k$ for $\alpha=192$.}
\end{figure}

\subsection{Visualization Results}
Visualization results for the main editing tasks and qualitative comparisons can be found in \texttt{Human\_Evaluation.pdf} included in the supplementary material.

\end{document}